%% file: main.tex
\documentclass{article} 
\usepackage{main,times}

\input{math_commands.tex}

\usepackage{hyperref}
\usepackage{url}

\usepackage[utf8]{inputenc} 
\usepackage[T1]{fontenc}    
\usepackage{hyperref}       
\usepackage{url}            
\usepackage{booktabs}       
\usepackage{amsfonts}       
\usepackage{nicefrac}       
\usepackage{microtype}      
\usepackage{xcolor}         
\usepackage{amsmath}

\usepackage{booktabs}
\usepackage{colortbl}
\usepackage{xcolor}
\usepackage{array}
\usepackage{graphicx}
\usepackage{multirow}

\usepackage{wrapfig}

\definecolor{oursrow}{RGB}{226, 239, 254}

\usepackage{colortbl}
\usepackage{array}

\definecolor{color3}{rgb}{0.95,0.95,0.95}
\definecolor{color4}{rgb}{0.90,0.9,0.9}

\title{%
\vspace{-2em}
  \hspace{-1em}%
  \raisebox{-2.5ex}{%
    \protect\includegraphics[height=5.0\fontcharht\font`\B]{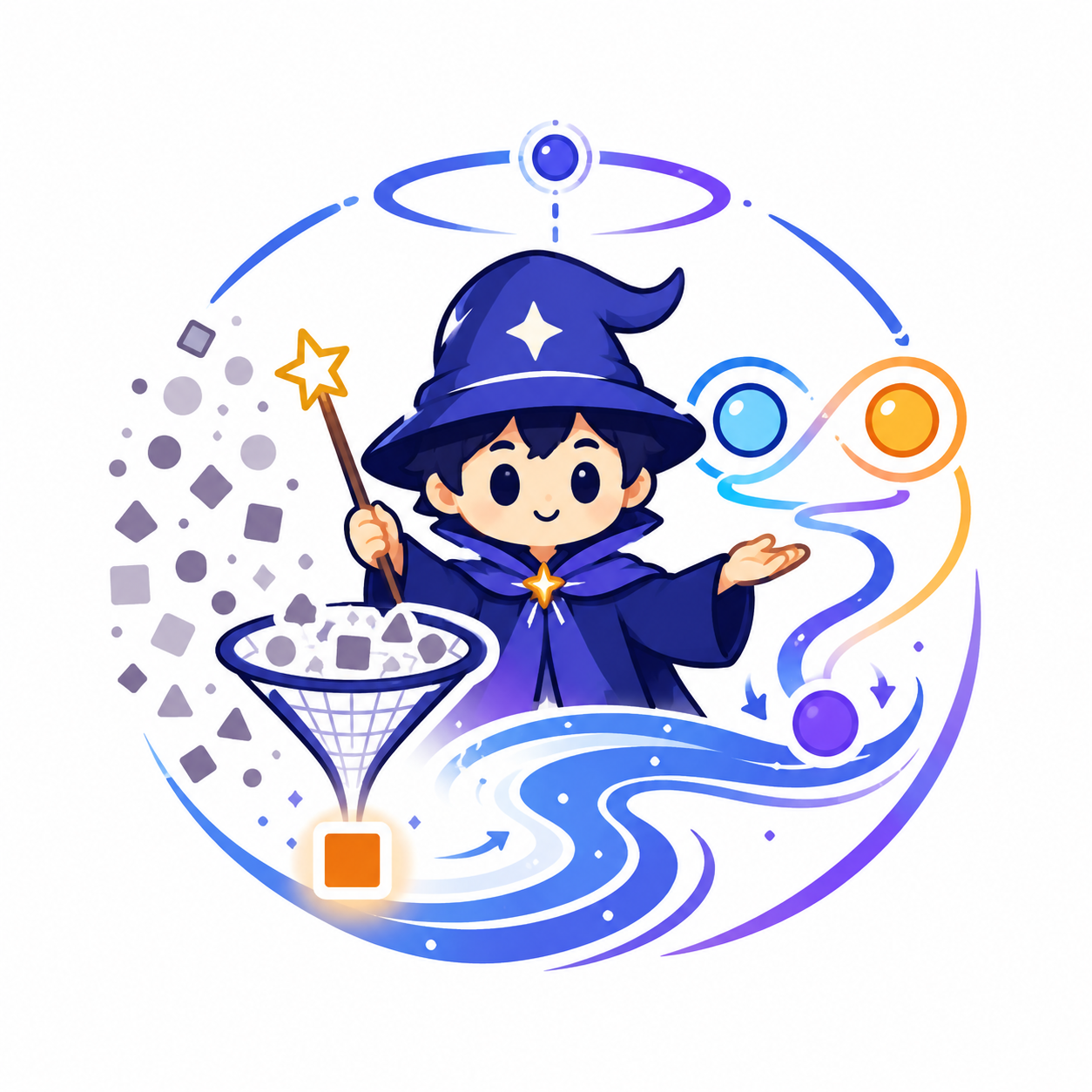}%
  }%
  AdaGRPO: A Capability-Aware Adaptive Enhancement for Flow-based GRPO
}

\iclrfinalcopy

\author{Jiazi Bu$^{1,2,3*}$\quad
Pengyang Ling$^{4*\S}$\quad
Yujie Zhou$^{1,3*}$\quad
Yibin Wang$^{8,6}$\quad
Yuhang Zang$^{3}$ \\
\textbf{Tianyi Wei$^{2}$}\quad
\textbf{Xiaohang Zhan$^{10}$}\quad
\textbf{Jiaqi Wang$^{6}$}\quad
\textbf{Tong Wu$^{5\dag}$}\quad
\textbf{Xingang Pan$^{2\dag}$}\quad
\textbf{Dahua Lin$^{3,7,9}$}\\
\small 
\textsuperscript{\rm 1}Shanghai Jiao Tong University \quad
\textsuperscript{\rm 2}S-Lab, Nanyang Technological University\quad
\textsuperscript{\rm 3}Shanghai AI Laboratory \\
\small
\textsuperscript{\rm 4}University of Science and Technology of China \quad
\textsuperscript{\rm 5}Stanford University \quad  
\textsuperscript{\rm 6}Shanghai Innovation Institute \\
\small
\textsuperscript{\rm 7}The Chinese University of Hong Kong \quad
\textsuperscript{\rm 8}Fudan University \quad
\textsuperscript{\rm 9}CPII under InnoHK \quad 
\textsuperscript{\rm 10}Adobe Research \\
\url{https://bujiazi.github.io/adagrpo.github.io/}
}

\begin{document}

{
  \renewcommand{\thefootnote}{\fnsymbol{footnote}}
  \footnotetext[1]{Equal contribution.  
  \textsuperscript{\S}Project leader.
  \textsuperscript{\dag}Corresponding author.}
}

\maketitle

\begin{figure}[h]
    \centering
    \vspace{-2em}
    \includegraphics[width=0.9\linewidth]{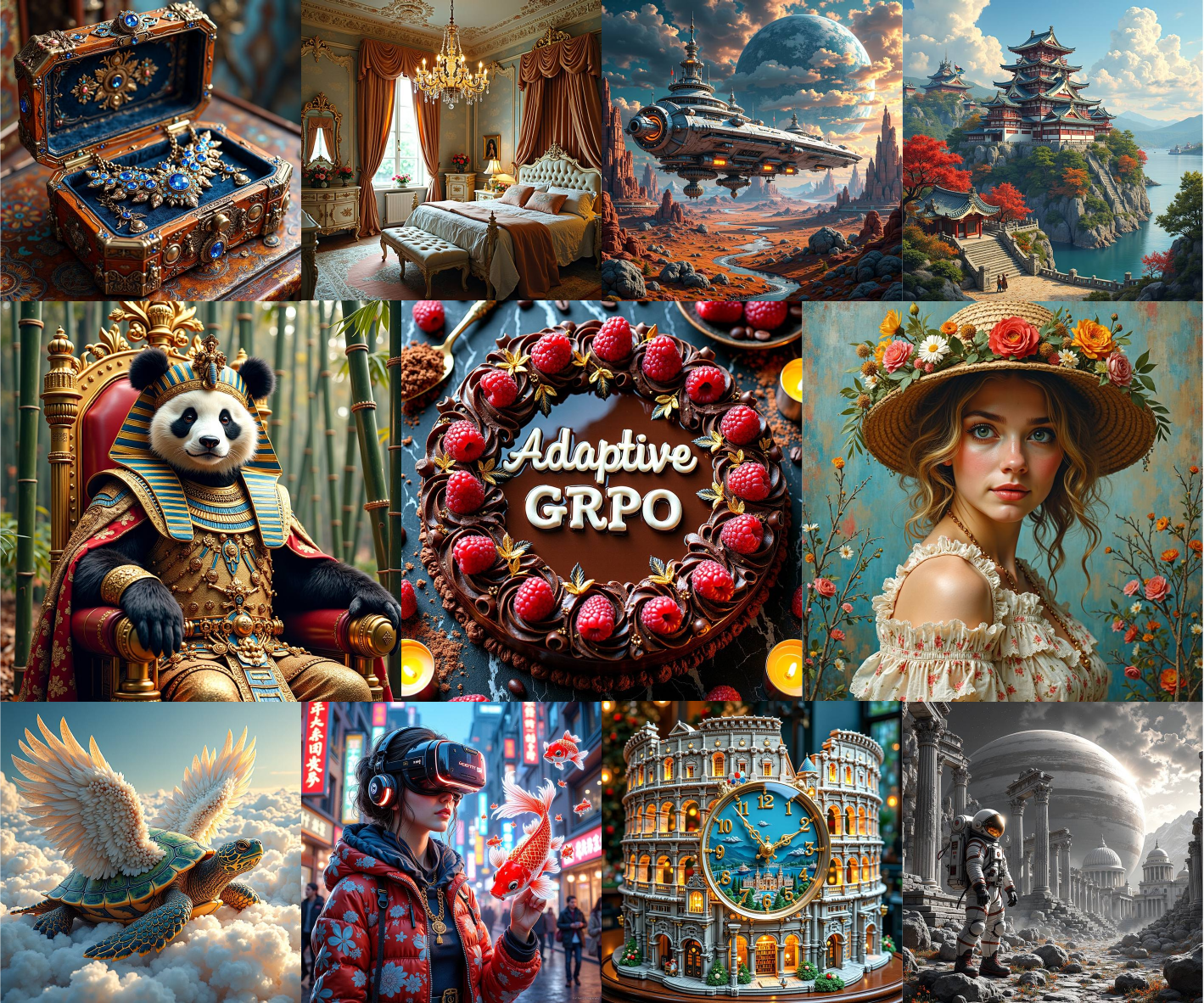}
    \vspace{-1em}
    \caption{
        \textbf{Gallery of AdaGRPO.}
        By integrating the proposed AdaGRPO, flow models (Flux.1-dev in this figure) experience a substantial leap in the generation performance, yielding remarkable improvements in intricate textures and visual fidelity. 
        \textbf{All prompts are listed in the appendix.}
        }
    \label{fig:teaser}
\end{figure}

\begin{abstract}
Group Relative Policy Optimization (GRPO) has demonstrated remarkable success in aligning text-to-image (T2I) flow models with human preferences. 
However, we have identified that the learning loop of current flow-based GRPO is fundamentally decoupled from the learner's current capability, suffering from critical blind spots at both \textit{prompt selection} and \textit{advantage estimation}: 
(i) existing methods sample prompts randomly, overlooking the substantial impact of data selection on reinforcement learning (RL) efficacy—a factor proven crucial in GRPO for large language models; 
and (ii) they evaluate sample quality solely relying on intra-group statistics, lacking a global perspective to accurately measure true policy improvement. 
To address these issues, we propose \textbf{Ada}ptive \textbf{GRPO} (\textbf{AdaGRPO}), a novel capability-aware RL algorithm tailored for flow models. 
Specifically, AdaGRPO consists of two principal components: (i) \textit{Online Curriculum Filtering Strategy} dynamically tracks the model's proficiency and adaptively selects prompts that best match its current learning boundary; 
(ii) \textit{Cross-Level Advantage Fusion} synergistically integrates fine-grained intra-group advantages with macro-level global advantages, providing a comprehensive and unbiased policy evaluation.
As a lightweight, plug-and-play module, AdaGRPO can be seamlessly integrated with existing frameworks such as Flow-GRPO, DanceGRPO, and Flow-CPS. Extensive experiments demonstrate that AdaGRPO consistently drives performance gains while significantly stabilizes GRPO training for flow models. 
\end{abstract}

\section{Introduction}

Recently, diffusion and flow-based models~\citep{dhariwal2021diffusion, ho2020denoising, podell2023sdxl, song2020denoising, song2020score} 
have firmly established themselves as the cornerstone of visual generation,
exhibiting remarkable proficiency in synthesizing high-quality visual contents~\citep{bu2025hiflow, flux2024, rombach2022high, hunyuanvideo2025, wan2025wan, Zhou_2025_ICCV}. 
Despite their impressive generation quality obtained through pre-training on large-scale datasets~\citep{schuhmann2022laion,nan2024openvid,chen2024panda},
these foundational models often suffer from misalignment with human preferences,
such as poor prompt adherence or aesthetic degradation.
Consequently, Reinforcement Learning from Human Feedback (RLHF)~\citep{black2023training, fan2023dpok, unifiedreward-flex} has become the popular approach for aligning T2I models. 
By leveraging reward models~\citep{kirstain2023pick, ma2025hpsv3, wang2026unified, wang2025unified, xu2023imagereward} explicitly designed to encapsulate human intent,
RL-based frameworks systematically steer the generation process toward favored visual characteristics and task-specific constraints.

Among various RL techniques~\citep{peng2025sudo, rafailov2023direct, schulman2017proximal, wallace2024diffusion},
Group Relative Policy Optimization (GRPO)~\citep{rafailov2023direct} has recently emerged as a highly promising alternative. 
By evaluating multiple generated samples for a given prompt and using intra-group comparison to estimate relative advantages,
GRPO bypasses the requirement of training a separate value network, 
making it well-suited for aligning large-scale models. 
To harness this potential for visual generation, 
an emerging line of research~\citep{liu2025flow, xue2025dancegrpo} has translated GRPO to flow models 
by replacing deterministic solvers with Stochastic Differential Equations (SDEs), 
thereby injecting the requisite exploration noise into the sampling trajectory.

Despite these successes, we posit that current flow-based GRPO frameworks~\citep{liu2025flow, xue2025dancegrpo, he2025tempflow, zhou2025g2rpo, li2025branchgrpo, li2025mixgrpo} 
are fundamentally decoupled from the model's evolving capability during training, suffering from blind spots at two foundational pillars of RL: \textit{prompt selection} (``what to learn from'') and \textit{advantage estimation} 
(``how to assign credit'').

\begin{figure}[t]
    \centering
    \includegraphics[width=1.0\linewidth]{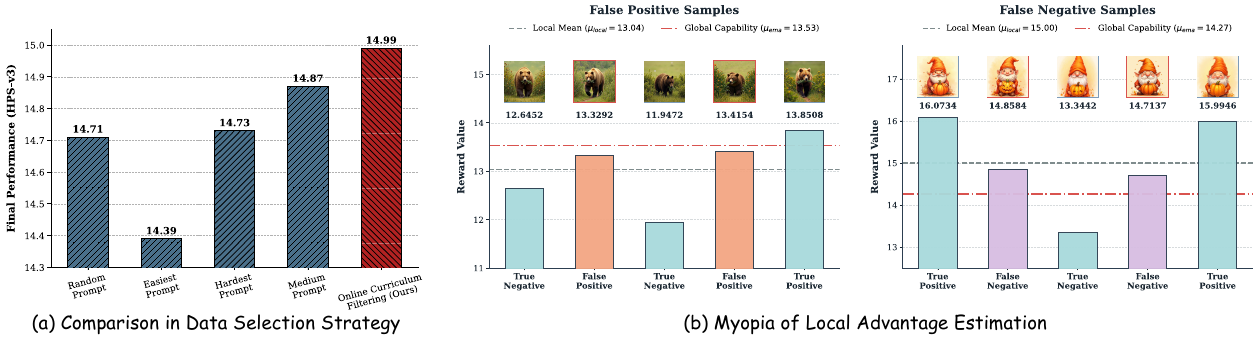}
    \vspace{-2.5em}
    \caption{
    \textbf{Observations.} 
    (a) Random sampling (current GRPO methods) and extreme prompts (``Easiest''/``Hardest'') yield suboptimal alignment efficacy. While selecting locally moderate prompts (``Medium'') offers improvements, it remains biased by the current batch. In contrast, our \textit{Online Curriculum Filtering Strategy} maximizes performance by dynamically identifying moderate tasks through the model's global capability.
    (b) Relying solely on local intra-group means erroneously produces false positive and false negative advantages that deviate from the model's global capability.
}
    \vspace{-1.5em}
    \label{fig:observation}
\end{figure}

Specifically, regarding \textit{prompt selection}, existing methods sample prompts blindly at random. Inspired by the success of prompt selection strategies in reinforcement learning alignment of large language models (LLMs)~\citep{zhang2025srpo,yu2025dapo}, we investigate the impact of prompt difficulty on flow-based GRPO. 
Prior to each training step, we profile all prompts in a candidate batch via their deterministic ODE rewards, then apply filtering heuristics to select which ones actually enter training.
As illustrated in Fig.~\ref{fig:observation} (a), training upon the ``easiest'' prompts (those yielding the highest rewards) causes severe performance degradation, while employing the ``hardest'' prompts (the lowest rewards) barely outperforms the random baseline. 
In contrast, prompts of medium difficulty drive notable gains, corroborating the established finding in LLM alignment that samples of moderate difficulty provide the most useful learning signal~\citep{bae2026online,cui2025process}.
However, the median reward of an isolated candidate batch is intrinsically biased, as it is susceptible to divergence from the model's aggregate proficiency. For instance, when an entire batch consists of challenging prompts, the median still exceeds the model's capability.
The absence of this global perspective also plagues \textit{advantage estimation}. 
Current methods typically evaluate samples solely via intra-group rewards and thus exhibit severe ``myopia''. In particular,
they erroneously assign positive advantages to subpar samples simply because they are above the local intra-group mean, 
even if they fall below the model's global capability (false positives)
, while penalize high-quality samples that fall below the local mean but actually surpass the global capability (false negatives), as shown in Fig.~\ref{fig:observation} (b). 
Without a reliable reference to gauge absolute policy progression, these local biases inevitably obscure the true optimization direction.

To this end, we propose \textbf{Ada}ptive \textbf{GRPO} (\textbf{AdaGRPO}), a novel capability-aware RL algorithm tailored for flow models, addressing the aforementioned blind spots through two principal components. 
First, \textit{Online Curriculum Filtering Strategy} is introduced to apply prompt selection. 
Rooted in curriculum learning~\citep{soviany2022curriculum}, 
this module maintains an Exponential Moving Average (EMA) of historical rewards to explicitly track the model's global generation proficiency, 
adaptively selecting candidate prompts perfectly at the current learning boundary.
This eliminates localized batch bias and ensures a highly constructive optimization landscape.
Second, \textit{Cross-Level Advantage Fusion} is proposed to calibrate advantage estimation. 
By synergistically fusing intra-group local advantages with macro-level global advantages, samples are rewarded not only for outperforming their immediate peers but also for surpassing the model's past capability bounds, yielding an unbiased signal of absolute policy progression.
As a lightweight, plug-and-play module, 
AdaGRPO seamlessly integrates into prevailing flow-based GRPO frameworks 
like Flow-GRPO~\citep{liu2025flow}, DanceGRPO~\citep{xue2025dancegrpo} and Flow-CPS~\citep{wang2025coefficients}.
Extensive experiments demonstrate that our method consistently drives multi-metric performance gains while significantly stabilizing GRPO training.

Our contributions are three-fold:
(i) We identify the structural decoupling in GRPO for flow models,
revealing that blind prompt sampling and myopic advantage estimation are bottlenecks causing training instability and suboptimal alignment.
To our best knowledge, we are \textit{the first to explore data selection in flow-based GRPO};
(ii) We propose \textbf{AdaGRPO}, a novel capability-aware RL algorithm featuring Online Curriculum Filtering Strategy for dynamic data curation and Cross-Level Advantage Fusion for unbiased advantage estimation;
(iii) AdaGRPO can be seamlessly integrated into diverse existing frameworks, offering superior preference alignment and more stable training process.

\section{Related Work}

\textbf{Diffusion and Flow Models}.
Diffusion models~\citep{ho2020denoising,song2020denoising,song2020score,dhariwal2021diffusion} learn to reverse a gradual noising process,
enabling high-fidelity visual synthesis across images~\citep{rombach2022high,podell2023sdxl,flux2024}, videos~\citep{guo2023animatediff,chen2024videocrafter2,wan2025wan}, and other modalities~\citep{voleti2024sv3d}. 
Flow matching models~\citep{esser2024scaling,lipman2022flow,liu2022flow} directly learn a continuous-time velocity field along straight-line trajectories between noise and data distributions, 
offering improved stability and scalability. 
Leading models such as Stable Diffusion~\citep{rombach2022high,podell2023sdxl}, Flux~\citep{flux2024,flux-2-2025}, Qwen-Image~\citep{wu2025qwen}, CogVideoX~\citep{yang2024cogvideox}, HunyuanVideo~\citep{kong2024hunyuanvideo,hunyuanvideo2025}, WAN~\citep{wan2025wan}, and LongCat-Video~\citep{team2025longcat} have demonstrated remarkable capabilities in generating high-quality visual content.

\textbf{Alignment for Diffusion and Flow Models}. 
Aligning diffusion/flow models with human preferences has evolved from early
PPO-based policy gradients~\citep{black2023training,schulman2017proximal,xu2023imagereward}
and DPO variants~\citep{peng2025sudo,rafailov2023direct,wallace2024diffusion} toward more efficient online RL frameworks. 
In particular, Group Relative Policy Optimization (GRPO)~\citep{shao2024deepseekmath} leverages intra-group relative advantages without a value network, 
inspiring adaptations to visual generation. 
Flow-GRPO~\citep{liu2025flow} and DanceGRPO~\citep{xue2025dancegrpo} reformulate deterministic ODE sampling into equivalent SDE trajectories to enable stochastic exploration, establishing the foundational paradigm for flow-based GRPO. 
Building on it, Flow-CPS~\citep{wang2025coefficients} eliminates SDE-induced noise artifacts by strictly aligning noise injection with the flow-matching scheduler.
Subsequent efforts further refine this paradigm from complementary perspectives,
such as enhancing training efficiency~\citep{li2025mixgrpo, zheng2025diffusionnft}, 
refining credit assignment~\citep{li2025branchgrpo, fu2025dynamic, he2025tempflow, zhou2025g2rpo},
and enriching reward formulations~\citep{Pref-GRPO&UniGenBench, bu2026sparse}.
Despite these advances, existing methods remain largely oblivious to the dynamic capability of the model. By relying on blind prompt sampling and myopic intra-group advantages, this structural decoupling leads to high training instability and suboptimal alignment efficiency. 

\textbf{Data Selection in Reinforcement Learning}.
Curriculum learning has long been recognized as an effective strategy to stabilize RL by exposing agents to tasks of progressively increasing difficulty~\citep{bengio2009curriculum, narvekar2020curriculum}. Recent works automate this process by aligning task selection with the agent's evolving capability: ProCuRL~\citep{tzannetos2023proximal} formalizes the Zone of Proximal Development to maximize learning progress, while Self-Paced RL~\citep{klink2020self} casts sampling as KL-regularized variational inference. 
In the era of LLM alignment, dynamic sampling techniques~\citep{bae2026online,cui2025process,zhang2025srpo} further refine data curation. 
Recent efforts have leveraged scoring metrics or group reward dynamics to filter uninformative or zero-variance prompts~\citep{chen2025scale, yu2025dapo, zheng2025act}, or formulated online task selection via Bayesian inference and Markov modeling to track solving dynamics~\citep{shen2025bots, mao2026dynamics}. 
As the first work to explore data selection within flow-based GRPO, AdaGRPO inherits the philosophy of curriculum learning. By dynamically tracking the capability of the learner and strategically selecting prompts that reside closest to its current learning boundary, our method effectively smooths the optimization landscape, ensuring robust and stable training progress.

\section{AdaGRPO}

\vspace{-0.5em}
\subsection{Preliminaries}
\vspace{-0.5em}
\textbf{Flow Matching as a Sequential Decision Process}. Flow matching transports samples from a Gaussian prior to a data distribution via a learned velocity field, which can be formulated as a finite-horizon Markov Decision Process (MDP). Given a condition $\mathbf{c}$, the generation trajectory of a flow model is defined as $\Gamma = (\mathbf{s}_T, \mathbf{a}_T, \mathbf{s}_{T-1}, \mathbf{a}_{T-1}, \dots, \mathbf{s}_0, \mathbf{a}_0)$, where each state is $\mathbf{s}_t = (\boldsymbol{x}_t, t, \mathbf{c})$ starting from $\boldsymbol{x}_T \sim \mathcal{N}(\mathbf{0}, \mathbf{I})$, and the action $\mathbf{a}_t$ corresponds to the single step denoising
process with the policy $\pi_\theta$.
The state transitions follow a deterministic ordinary differential equation (ODE):
\begin{equation}
\frac{d\boldsymbol{x}_t}{dt} = \boldsymbol{v}_\theta(\boldsymbol{x}_t, t, \mathbf{c}).
\end{equation}
where $\boldsymbol{v}_\theta(\boldsymbol{x}_t, t, \mathbf{c})$ is the predicted velocity. While this deterministic mapping ensures high-fidelity generation, it inherently lacks the stochasticity required for RL exploration.

\noindent\textbf{ODE-to-SDE Conversion}. To adapt flow models for online reinforcement learning, prior works transform the deterministic ODE into an equivalent Stochastic Differential Equation (SDE). By introducing a diffusion term and compensating the drift, the dynamics become:
\begin{equation}
d\boldsymbol{x}_t = \left( \boldsymbol{v}_\theta(\boldsymbol{x}_t, t, \mathbf{c}) + \frac{\sigma_t^2}{2t} \big( \boldsymbol{x}_t + (1-t) \boldsymbol{v}_\theta(\boldsymbol{x}_t, t, \mathbf{c}) \big) \right) dt + \sigma_t d\mathbf{w}_t,
\end{equation}
where $\mathbf{w}_t$ is the standard Wiener process and $\sigma_t = \eta \sqrt{t / (1-t)}$ governs the magnitude of injected noise with a hyperparameter $\eta$.
Discretizing this via the Euler–Maruyama scheme over $\Delta t$ yields:
\begin{equation}
\boldsymbol{x}_{t+\Delta t} = \boldsymbol{x}_t + \left[ \boldsymbol{v}_\theta(\boldsymbol{x}_t, t, \mathbf{c}) + \frac{\sigma_t^2}{2t} \big( \boldsymbol{x}_t + (1-t) \boldsymbol{v}_\theta(\boldsymbol{x}_t, t, \mathbf{c}) \big) \right] \Delta t + \sigma_t \sqrt{\Delta t} \, \boldsymbol{\epsilon},
\end{equation}
with $\boldsymbol{\epsilon} \sim \mathcal{N}(\mathbf{0}, \mathbf{I})$. This stochastic formulation injects necessary variance for policy gradient estimation without altering the underlying generative distribution.

\noindent\textbf{GRPO Objective}. Group Relative Policy Optimization (GRPO) is a value-free RL paradigm that aligns policies using intra-group feedback. Given a prompt $\mathbf{c}$, the current policy $\pi_{\theta_{\text{old}}}$ samples $G$ trajectories via the SDE, yielding a group of terminal samples $\{\boldsymbol{x}_0^i\}_{i=1}^G$. The advantage for the $i$-th sample at any timestep $t$ is computed via group-wise reward normalization:
\begin{equation}
\hat{A}_t^i = \frac{R(\boldsymbol{x}_0^i, \mathbf{c}) - \text{mean}\big(\{R(\boldsymbol{x}_0^j, \mathbf{c})\}_{j=1}^G\big)}{\text{std}\big(\{R(\boldsymbol{x}_0^j, \mathbf{c})\}_{j=1}^G\big)}.
\end{equation}
The policy is then updated by maximizing a clipped surrogate objective with a KL penalty:
\begin{equation}\label{eq:objective}
\begin{aligned}
\mathcal{J}_{\text{GRPO}}(\theta) = \mathbb{E}_{\mathbf{c}, \{\boldsymbol{x}^i\}} \Bigg[ \frac{1}{G} \sum_{i=1}^G \frac{1}{T} \sum_{t=0}^{T-1} \bigg(
&\min\big(r_t^i(\theta) \hat{A}_t^i, \text{clip}(r_t^i(\theta), 1-\varepsilon, 1+\varepsilon) \hat{A}_t^i\big) \\
&- \beta D_{\text{KL}}\big(\pi_\theta(\cdot|\boldsymbol{x}_t,\mathbf{c}) \,\|\, \pi_{\text{ref}}(\cdot|\boldsymbol{x}_t,\mathbf{c})\big)
\bigg) \Bigg],
\end{aligned}
\end{equation}
where $r_t^i(\theta) = \pi_\theta(\boldsymbol{x}_{t-1}^i | \boldsymbol{x}_t^i, \mathbf{c}) / \pi_{\theta_{\text{old}}}(\boldsymbol{x}_{t-1}^i | \boldsymbol{x}_t^i, \mathbf{c})$ is the importance sampling ratio, $\varepsilon$ is the clip threshold, $\beta$ weights the KL penalty against the reference policy $\pi_{\text{ref}}$.

\vspace{-0.5em}
\subsection{Online Curriculum Filtering Strategy}\label{sec:curriculum_filtering}
\vspace{-0.5em}

Existing flow-based GRPO methods~\citep{liu2025flow,xue2025dancegrpo} sample training prompts uniformly at random. 
This blind strategy frequently exposes the policy to extreme tasks that yield either noisy or uninformative optimization signals~\citep{cui2025process,bae2026online}. Furthermore, while applying a localized filtering heuristic (e.g., selecting the median prompt within a candidate batch, as shown in Fig.~\ref{fig:observation} (a)) can alleviate this issue, it remains biased by the current batch distribution and is disconnected from the model's dynamically evolving capability.

To overcome this, we propose the \textit{Online Curriculum Filtering Strategy}, a lightweight yet effective mechanism rooted in the philosophy of curriculum learning~\citep{soviany2022curriculum}. Instead of relying on restricted local heuristics, the core idea is to continuously track the model's global generation proficiency and adaptively select candidate prompts that reside perfectly at its current learning boundary. Such genuinely moderate prompts consistently induce the constructive reward variance necessary for meaningful intra-group ranking, providing a clear optimization direction.

Specifically, at each training iteration $k$, instead of directly performing the SDE group rollout on a random prompt, we first sample a small batch of candidate prompts $\mathcal{B} = \{\mathbf{c}_1, \mathbf{c}_2, \dots, \mathbf{c}_B\}$. For each prompt $\mathbf{c}_b$, we perform a single deterministic ODE sampling to generate a baseline sample $\boldsymbol{x}_0^{b,\text{ODE}} \sim \pi_{\theta_{\text{old}}}(\cdot | \mathbf{c}_b)$, and compute its corresponding reward $R_b^\text{ODE} = R(\boldsymbol{x}_0^{b,\text{ODE}}, \mathbf{c}_b)$.
To establish a stable capability anchor, we maintain a global historical reward baseline using an Exponential Moving Average (EMA). Let $\mu_{\text{ema}}^{(k)}$ denote the historical mean reward up to iteration $k$, we update this capability anchor using the mean ODE reward of the current candidate batch:
\begin{equation}
    \mu_{\text{ema}}^{(k)} = \alpha \mu_{\text{ema}}^{(k-1)} + (1 - \alpha) \frac{1}{B} \sum_{b=1}^B R_b^\text{ODE},
\end{equation}
where $\alpha \in (0, 1)$ is the momentum coefficient. Then, we track the EMA variance $(\sigma_{\text{ema}}^{(k)})^2$ to capture the global reward distribution, which is used in \textit{Cross-Level Advantage Fusion} module (Section \ref{sec:advantage_estimation}):
\begin{equation}
    (\sigma_{\text{ema}}^{(k)})^2 = \alpha (\sigma_{\text{ema}}^{(k-1)})^2 + (1 - \alpha) \frac{1}{B} \sum_{b=1}^B \left( R_b^\text{ODE} - \mu_{\text{ema}}^{(k)} \right)^2.
\end{equation}
The EMA mean $\mu_{\text{ema}}^{(k)}$ serves as a robust proxy for the model's current generation capability. For standard single-reward optimization, the candidate batch is filtered to select the prompt $\mathbf{c}_{b^*}$ whose ODE reward is closest to the current capability anchor:
\begin{equation}
    b^* = \arg\min_{b \in \{1, \dots, B\}} \left| R_b^\text{ODE} - \mu_{\text{ema}}^{(k)} \right|.
\end{equation}

Furthermore, this strategy can be seamlessly extended to joint multi-reward optimization involving $M$ distinct reward models, where the optimal prompt is identified by minimizing the sum of normalized deviations across all reward signals:
\begin{equation}
    b^* = \arg\min_{b \in \{1, \dots, B\}} \sum_{m=1}^M \frac{\left| R_{b, m}^\text{ODE} - \mu_{\text{ema}, m}^{(k)} \right|}{\mu_{\text{ema}, m}^{(k)}},
\end{equation}
where $R_{b,m}^\text{ODE}$ and $\mu_{\text{ema},m}^{(k)}$ denote the $m$-th reward value and its corresponding capability anchor, respectively. The normalization operation eliminates the scale discrepancies among different reward models.
Once the optimal prompt $\mathbf{c}_{b^*}$ is selected, we execute the full stochastic group rollout (size $G$) exclusively on $\mathbf{c}_{b^*}$ for subsequent optimization.

\begin{figure}[t]
    \centering
    \includegraphics[width=1.0\linewidth]{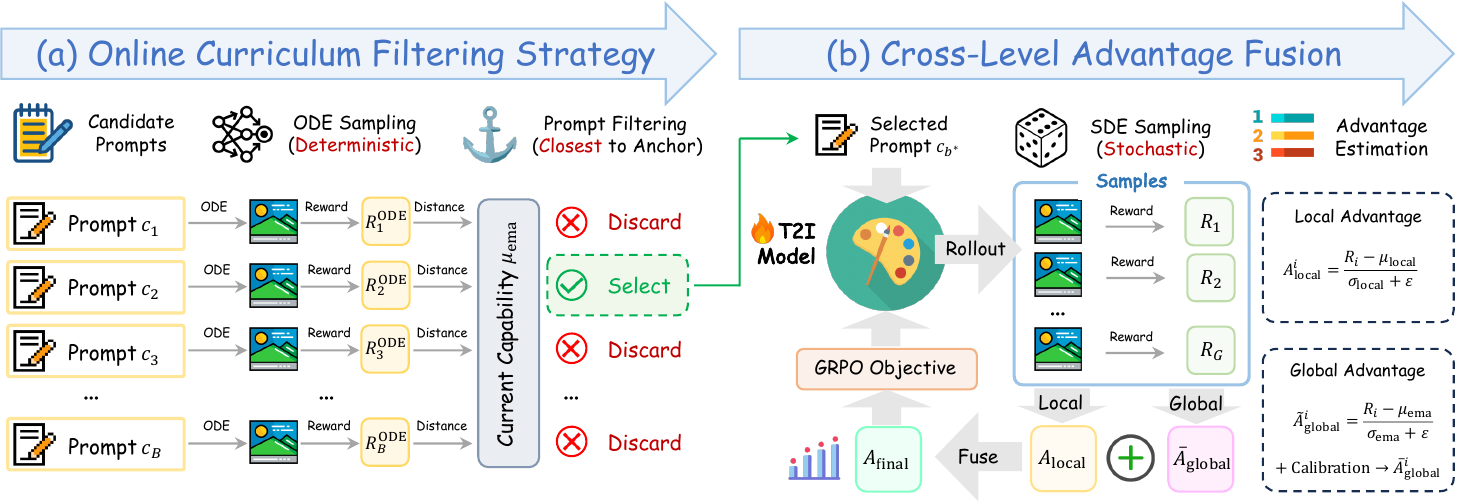}
    \vspace{-2.5em}
    \caption{
    \textbf{Pipeline of AdaGRPO}. (a) First, \textit{Online Curriculum Filtering Strategy} evaluates candidate prompts via deterministic ODE sampling and adaptively selects the one that best matches the model's current capability anchor ($\mu_{\text{ema}}$). 
    The selected prompt is utilized for stochastic SDE rollout. (b) Then, \textit{Cross-Level Advantage Fusion} integrates the intra-group local advantage with the history-calibrated global advantage to formulate an unbiased, comprehensive signal ($A_{\text{final}}$) for GRPO optimization.
}
    \vspace{-1em}
    \label{fig:pipeline}
\end{figure}

\vspace{-0.5em}
\subsection{Cross-Level Advantage Fusion}\label{sec:advantage_estimation}
While the standard GRPO effectively avoids the need for a separate value network, its exclusive reliance on intra-group evaluation restricts its field of view. As demonstrated in Fig.~\ref{fig:observation} (b), this localized paradigm generates false positive advantages by rewarding subpar samples simply because they exceed the local batch mean, while producing false negative advantages by penalizing high-quality samples that happen to fall below their immediate peers but actually surpass the global capability. Such miscalibrated evaluations obscure genuine absolute policy progression. To this end, we propose \textit{Cross-Level Advantage Fusion}, which synergistically integrates fine-grained local rankings with a macro-level global capability baseline.

\textbf{Local Advantage Estimation.} Given the selected prompt $\mathbf{c}_{b^*}$ and its generated group of $G$ samples $\{\boldsymbol{x}_0^i\}_{i=1}^G$, we first compute the standard intra-group local advantage (timestep $t$ is omitted for brevity):
\begin{equation}
    A_{\text{local}}^i = \frac{R_i - \mu_{\text{local}}}{\sigma_{\text{local}} + \epsilon},
\end{equation}
where $R_i$ is the reward of the $i$-th SDE sample generated with $\mathbf{c}_{b^*}$, $\mu_{\text{local}}$ and $\sigma_{\text{local}}$ are the mean and standard deviation of $\{R_i\}_{i=1}^G$, respectively, and $\epsilon$ is a small constant for numerical stability.

\textbf{Global Advantage Calibration.} To inject a global perspective, we leverage the historical reward statistics, $\mu_{\text{ema}}^{(k)}$ and $\sigma_{\text{ema}}^{(k)}$, maintained in Section \ref{sec:curriculum_filtering}, to derive a raw global advantage:
\begin{equation}
    \tilde{A}_{\text{global}}^i = \frac{R_i - \mu_{\text{ema}}^{(k)}}{\sigma_{\text{ema}}^{(k)} + \epsilon}.
\end{equation}
However, directly utilizing $\tilde{A}_{\text{global}}^i$ breaks the critical zero-mean property of GRPO, potentially destabilizing the policy optimization. To enforce a strict zero-mean distribution while preserving the sign of each sample's absolute progression, we introduce a conditional sign-preserving normalization step. Let $\mathcal{P} = \{i \mid \tilde{A}_{\text{global}}^i > 0\}$ and $\mathcal{N} = \{i \mid \tilde{A}_{\text{global}}^i < 0\}$ denote the indices of positive and negative global advantages, respectively. They are scaled conditionally as follows:
\begin{equation}
    \bar{A}_{\text{global}}^i =
    \begin{cases}
        \frac{\tilde{A}_{\text{global}}^i}{\sum_{j \in \mathcal{P}} \tilde{A}_{\text{global}}^j}, & \text{if } i \in \mathcal{P} \text{ and } \mathcal{P}, \mathcal{N} \neq \emptyset \\
        \frac{\tilde{A}_{\text{global}}^i}{\sum_{j \in \mathcal{N}} |\tilde{A}_{\text{global}}^j|}, & \text{if } i \in \mathcal{N} \text{ and } \mathcal{P}, \mathcal{N} \neq \emptyset \\
        \tilde{A}_{\text{global}}^i - \frac{1}{G}\sum_{j=1}^G \tilde{A}_{\text{global}}^j, & \text{otherwise.}
    \end{cases}
\end{equation}
In the standard case (both sets non-empty), this operation scales the sum of positive terms to $1$ and negative terms to $-1$, guaranteeing $\sum_{i} \bar{A}_{\text{global}}^i = 0$. In the rare event of a unilateral batch (i.e., $\mathcal{P} = \emptyset$ or $\mathcal{N} = \emptyset$), we dynamically fall back to standard mean-centering to prioritize training stability.

\textbf{Cross-Level Fusion.} Finally, we formulate the comprehensive advantage signal by aggregating the local and global advantages:
\begin{equation}
    A_{\text{final}}^i = A_{\text{local}}^i + \bar{A}_{\text{global}}^i.
\end{equation}
By replacing the standard advantage $\hat{A}_t^i$ in Equation~\ref{eq:objective} with this fused advantage $A_{\text{final}}^i$, AdaGRPO ensures that samples are rewarded not only for outperforming their peers but also for surpassing the model's historical capability bounds, providing an unbiased gradient direction.  The overview of our AdaGRPO framework is illustrated in Fig.~\ref{fig:pipeline}.

\begin{figure}[t]
    \centering
    \includegraphics[width=1.0\linewidth]{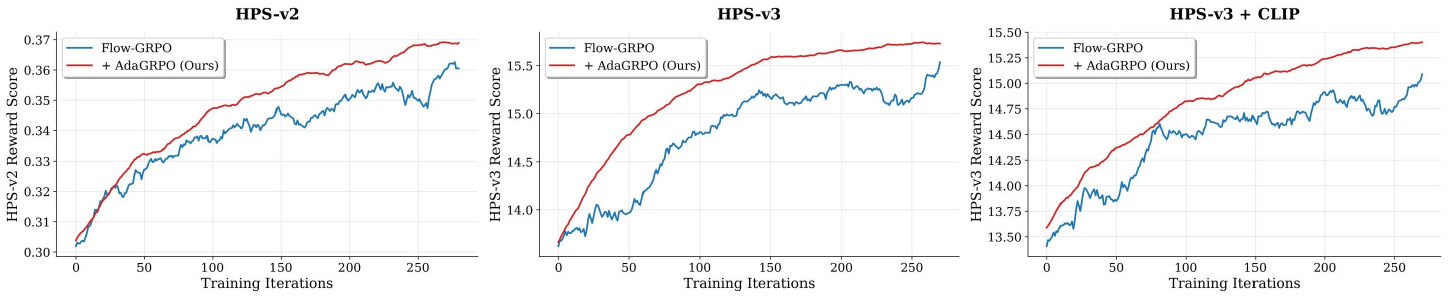}
    \vspace{-2.5em}
    \caption{
        \textbf{Reward Curves during Training.} The proposed AdaGRPO facilitates significantly smoother training dynamics and higher performance ceilings across diverse training configurations.
        }
    \vspace{-1.5em}
    \label{fig:curve}
\end{figure}

\section{Experiments}
\vspace{-0.5em}
\subsection{Implementation Details}\label{subsec:implementation_details}
\textbf{Datasets and Models}. Following prior works~\citep{xue2025dancegrpo, li2025mixgrpo, zhou2025g2rpo}, the HPD~\citep{wu2023human} dataset is utilized as the prompt dataset. 
This comprehensive corpus supplies over 100K diverse prompts to drive the RL training, alongside a distinct set of 400 prompts for evaluation.
For the generative backbone, all our experiments are built upon Flux.1-dev~\citep{flux2024}, one of the most capable open-sourced flow models currently available. 
Further implementation details are provided in Section~\ref{sec:add-implementation} in the appendix.

\textbf{Baselines}. To demonstrate its architecture-agnostic nature, we implement AdaGRPO upon three representative flow-based GRPO baselines: Flow-GRPO~\citep{liu2025flow}, DanceGRPO~\citep{xue2025dancegrpo}, and Flow-CPS~\citep{wang2025coefficients}. To improve training efficiency, the few-step training mechanism of Flow-GRPO-Fast~\citep{liu2025flow} is adopted by all assessed methods.

\textbf{Evaluation Metrics}. For a comprehensive assessment, we assemble a diverse suite of automated evaluation metrics that capture different facets of generation quality, including (i) \textit{CLIP/BLIP-based Reward Models}: HPS-v2~\citep{wu2023human}, CLIP~\citep{radford2021learning} and ImageReward (\textbf{IR})~\citep{xu2023imagereward}; (ii) \textit{LVLM-based Reward Models}: HPS-v3~\citep{ma2025hpsv3} and UnifiedReward-v1/v2 (\textbf{UR-v1/v2})~\citep{wang2025unified}; and (iii) \textit{General T2I Benchmarks}: UniGenBench~\citep{wang2025unigenbench++}, a unified and versatile benchmark for image generation. This comprehensive benchmark covers ten distinct categories that span essential aspects such as conceptual fidelity, visual appeal, and text-image alignment, offering a holistic measure of generative capability.

\textbf{Training Paradigms}. Following previous studies~\citep{xue2025dancegrpo, li2025mixgrpo, zhou2025g2rpo}, we train AdaGRPO under two distinct training configurations: 
(i) \textit{Single Reward}: the flow model is fine-tuned using a solitary reward signal (specifically, either HPS-v2 or HPS-v3); (ii) \textit{Multi-Reward}: the policy is jointly optimized under signals from both HPS-v3 and CLIP for more robust and generalizable alignment outcomes. 
The training results of our AdaGRPO on UnifiedReward-v2 are presented in Section~\ref{sec:add-quantitative} in the appendix.

\textbf{Sampling Details}. Following previous works~\citep{xue2025dancegrpo, zhou2025g2rpo}, a group size of $G=12$ is adopted and the total number of sampling steps is configured as $T=16$. 
The candidate prompt batch size $B$ and the momentum coefficient $\alpha$ are set to $10$ and $0.6$, respectively.

\textbf{Optimization Details}. All experiments are produced on $8\times$ NVIDIA H200 GPUs with the batch size setting to $1$. The AdamW optimizer is utilized with a learning rate of $2\times10^{-6}$ and a weight decay of $1\times10^{-4}$. 
For efficiency, \texttt{bfloat16} (bf16) mixed-precision is leveraged during training. 

\begin{figure}[t]
    \centering
    \includegraphics[width=1.0\linewidth]{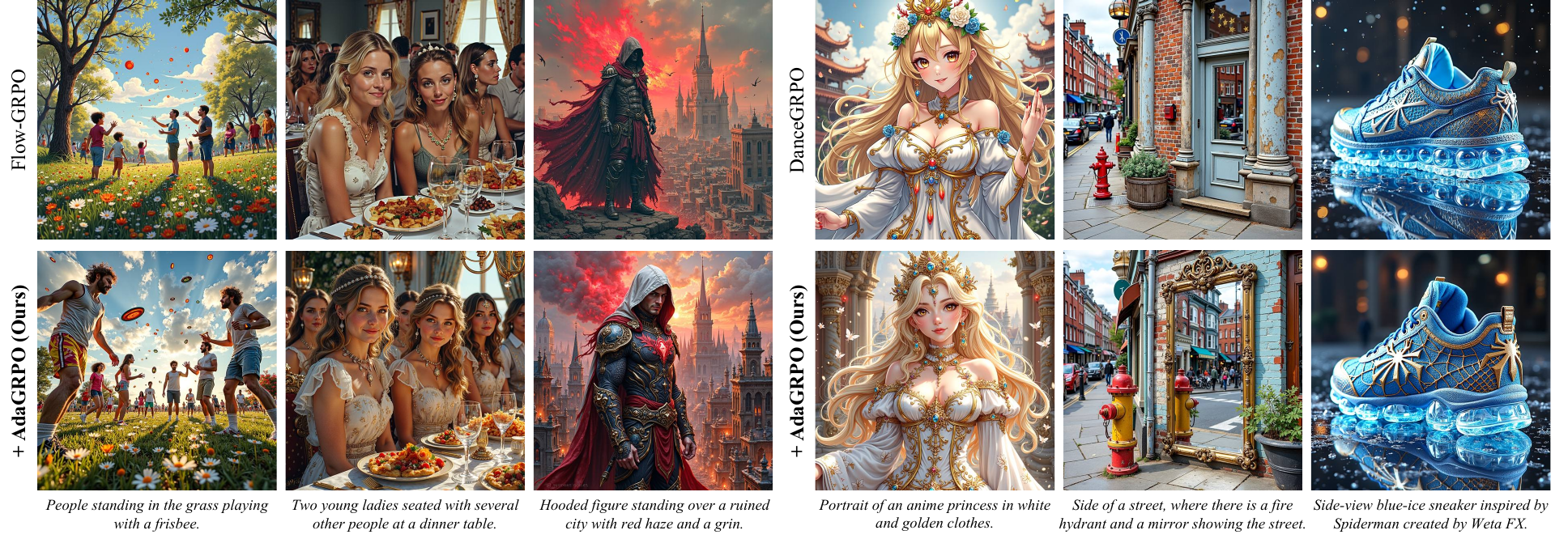}
    \vspace{-2.5em}
    \caption{
        \textbf{Qualitative Comparisons with Baselines on HPS-v2.} Best viewed zoomed in.
        }
    \label{fig:hpsv2}
    \vspace{-2em}
\end{figure}

\input{tabs/quant}

\subsection{Main Results}
\textbf{Quantitative Evaluation}. The quantitative assessments are presented in Tab.~\ref{tab:comparison} and Tab.~\ref{tab:unigenbench}. Under both single reward (HPS-v2/v3) and multi-reward (HPS-v3 + CLIP) settings, AdaGRPO consistently brings substantial improvements to the prevailing baselines (Flow-GRPO, DanceGRPO, and Flow-CPS), validating its effectiveness and architecture-agnostic nature. Specifically, as shown in Tab.~\ref{tab:comparison}, AdaGRPO delivers the best performance on the majority of evaluation metrics, with particularly notable gains in HPS-related scores, coherence (UR-v2-C), style (UR-v2-S) and ImageReward. 
Meanwhile, when incorporating the CLIP reward model to enforce semantic alignment during joint multi-reward training, AdaGRPO achieves consistent improvements across nearly all evaluation dimensions.
Furthermore, as detailed in Tab.~\ref{tab:unigenbench}, the comprehensive evaluation on UniGenBench corroborates our superiority in fine-grained visual synthesis. 
The training reward curves for Flow-GRPO (with or without AdaGRPO) are presented in Fig.~\ref{fig:curve}.

\begin{figure}[t]
    \centering
    \includegraphics[width=1.0\linewidth]{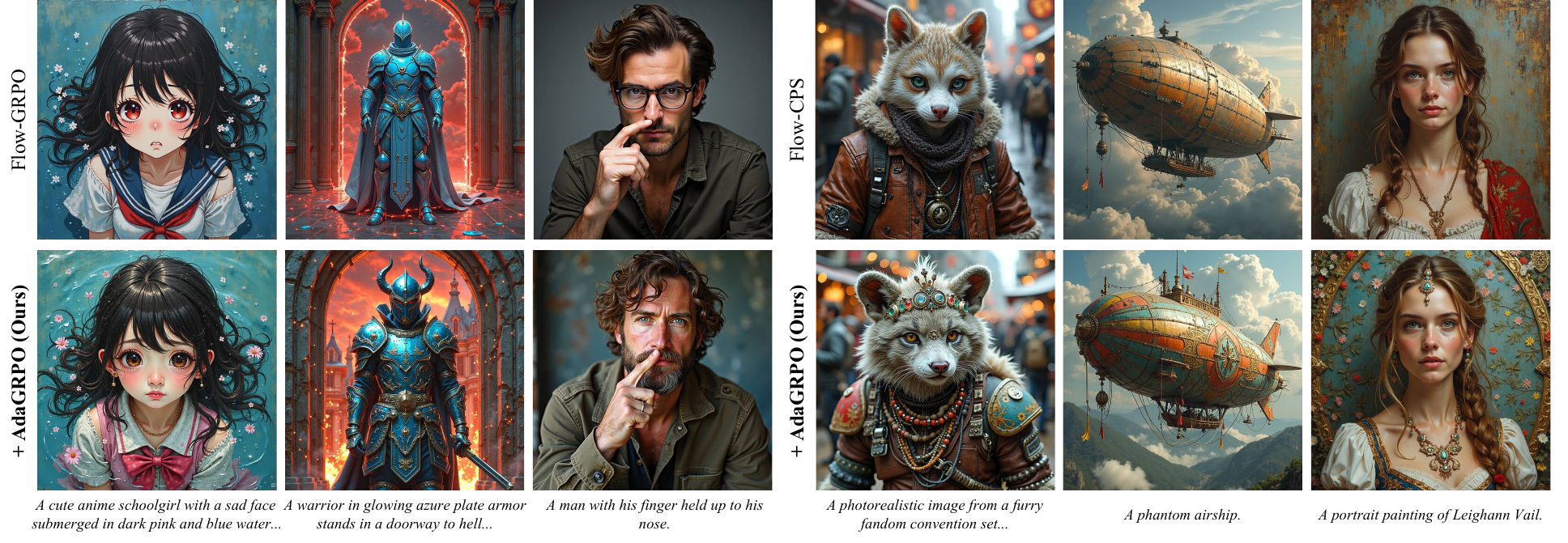}
    \vspace{-2.5em}
    \caption{
        \textbf{Qualitative Comparisons with Baselines on HPS-v3.} Best viewed zoomed in.
        }
    \label{fig:hpsv3}
    \vspace{-1.5em}
\end{figure}

\textbf{Qualitative Comparison}. As depicted in Fig.~\ref{fig:hpsv2} and Fig.~\ref{fig:hpsv3}, AdaGRPO consistently surpasses the baselines in visual fidelity, aesthetic appeal, and semantic adherence. In the \textit{``dinner table''} and \textit{``man finger nose''} cases, our method renders portraits with significantly more natural skin textures, precise facial anatomy, and cinematic lighting gradients, overcoming the plastic appearance generated by baselines. For the \textit{``warrior''} and \textit{``blue-ice sneaker''} examples, AdaGRPO substantially enhances the visual richness by synthesizing intricate ornamental details, vivid glowing elements, and realistic material reflections. Furthermore, our method exhibits robust adherence to complex spatial compositions. In the \textit{``mirror''} case, the baseline completely ignores the specified reflective element, whereas AdaGRPO accurately generates an ornate mirror reflecting the street scene. Similarly, in the \textit{``frisbee''} case, it transforms a static and chaotic baseline generation into a highly dynamic action composition with a more immersive atmosphere. More results are provided in Section~\ref{sec:add-qualitative} in appendix.

\input{tabs/unigen}

\subsection{Ablation and Analysis}
We conducted ablation studies of AdaGRPO on Flow-GRPO framework under the HPS-v2 training setting. Given that the two proposed components build upon one another, we begin by ablating the Online Curriculum Filtering Strategy, and subsequently investigate the impact of incorporating the Cross-Level Advantage Fusion under its optimal configuration. The results are presented in Tab.~\ref{tab:ablation}.

\textbf{Effects of Online Curriculum Filtering Strategy}. 
This strategy relies on two hyperparameters: the momentum coefficient $\alpha$ and the candidate batch size $B$. 
$\alpha$ controls the update rate of the historical capability anchor. As depicted in Tab.~\ref{tab:ablation} (a), a moderate $\alpha=0.6$ achieves the best performance by effectively balancing long-term historical knowledge with current batch statistics. 
Meanwhile, $B$ defines the search space for prompt selection. While a larger $B$ theoretically enables more precise capability matching, Tab.~\ref{tab:ablation} (b) reveals a clear diminishing return when scaling $B$ beyond 10. Considering the computational overhead of additional ODE sampling, $B=10$ is chosen to strike a balance between training efficiency and alignment performance.

\input{tabs/ablation_right}

\textbf{Effects of Cross-Level Advantage Fusion}. 
As shown in Tab.~\ref{tab:ablation} (c), relying solely on intra-group evaluation traps the policy in local optima, yielding suboptimal performance. By contrast, integrating the historical baseline significantly elevates both HPS-related scores and the averaged UR-v2 metrics, confirming that our cross-level fusion effectively rectifies myopic local biases and drives genuine policy progression.

\section{Conclusion}
\vspace{-0.5em}
In this paper, we identify that blind prompt selection and myopic advantage estimation in current flow-based GRPO lead to training instability and suboptimal alignment. To address this, we propose AdaGRPO, a lightweight capability-aware RL framework. It features \textit{Online Curriculum Filtering Strategy} to dynamically match training prompts with the model's evolving capability, and \textit{Cross-Level Advantage Fusion} to integrate local rankings with a global baseline for unbiased policy evaluation. Extensive experiments demonstrate that AdaGRPO seamlessly integrates into prevailing architectures, consistently driving superior generation quality and highly stable training dynamics.

\newpage
\bibliographystyle{main}
\bibliography{main}

\newpage
\appendix

\section{Appendix}
In the appendix, we present additional implementation details (Section~\ref{sec:add-implementation}), additional quantitative results (Section~\ref{sec:add-quantitative}), additional qualitative results (Section~\ref{sec:add-qualitative}), text prompts for image generation in both the main paper and appendix (Section~\ref{sec:prompt}), the limitation of our method (Section~\ref{sec:add-limitation}), the ethical statement (Section~\ref{sec:add-ethical}), the reproducibility statement (Section~\ref{sec:add-reproducibility}), as well as the declaration on LLM usage (Section~\ref{sec:add-llm}), as a supplement to the main paper. 

\section{Additional Implementation Details}
\label{sec:add-implementation}

Tab.~\ref{tab:hyperparams} presents the detailed hyperparameter settings used in our study, which were kept consistent throughout all experiments.

\input{tabs/hyperparameters}

\section{Additional Quantitative Results}
\label{sec:add-quantitative}

To further validate the versatility of our proposed AdaGRPO, we conduct additional experiments utilizing UnifiedReward-v2 (UR-v2)~\citep{wang2025unified} as the reward model.
Unlike CLIP or HPS variants, UR-v2 is a state-of-the-art LVLM-based reward model that provides comprehensive, multi-dimensional evaluations encompassing Alignment (UR-v2-A), Coherence (UR-v2-C), and Style (UR-v2-S).
Specifically, we integrate AdaGRPO into the three representative baseline frameworks (Flow-GRPO, DanceGRPO, and Flow-CPS) and train them using the averaged UR-v2 score.
As shown in Tab.~\ref{tab:urv2_quant}, AdaGRPO consistently outperforms the standard GRPO baselines across all architectures, achieving superior scores on the target UR-v2 dimensions while exhibiting robust generalization to unseen auxiliary metrics such as HPS-v2/v3, UR-v1, and ImageReward (IR).

\input{tabs/unifiedrewardv2}

\section{Additional Qualitative Results}
\label{sec:add-qualitative}

\begin{figure}[h]
    \centering
    \includegraphics[width=1.0\linewidth]{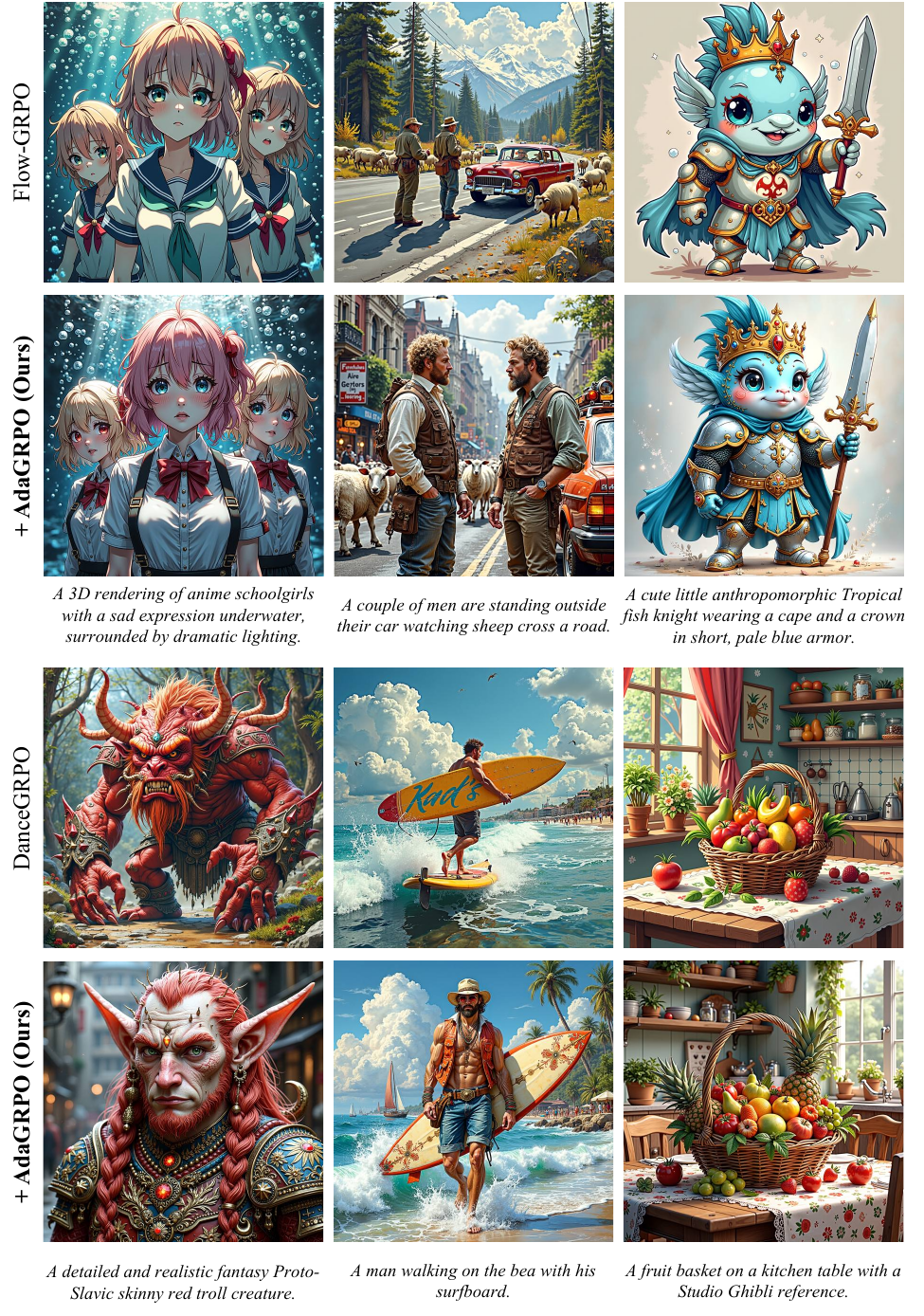}
    \vspace{-1em}
    \caption{
        \textbf{Additional Comparison Results on HPS-v2. (1/2)}
        }
    \label{fig:hpsv2-1}
\end{figure}

\begin{figure}[h]
    \centering
    \includegraphics[width=1.0\linewidth]{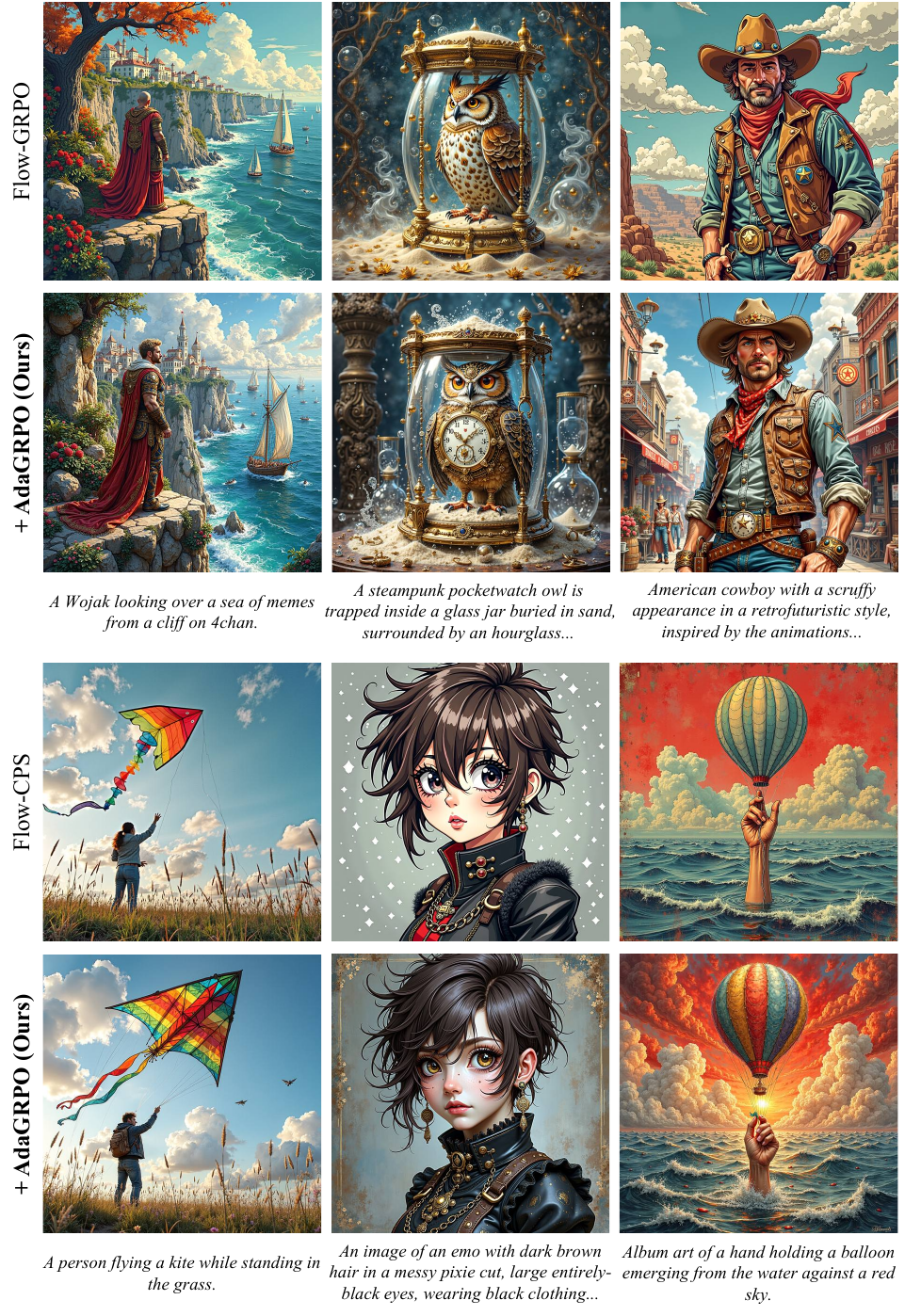}
    \vspace{-1em}
    \caption{
        \textbf{Additional Comparison Results on HPS-v2. (2/2)}
        }
    \label{fig:hpsv2-2}
\end{figure}

\begin{figure}[h]
    \centering
    \includegraphics[width=1.0\linewidth]{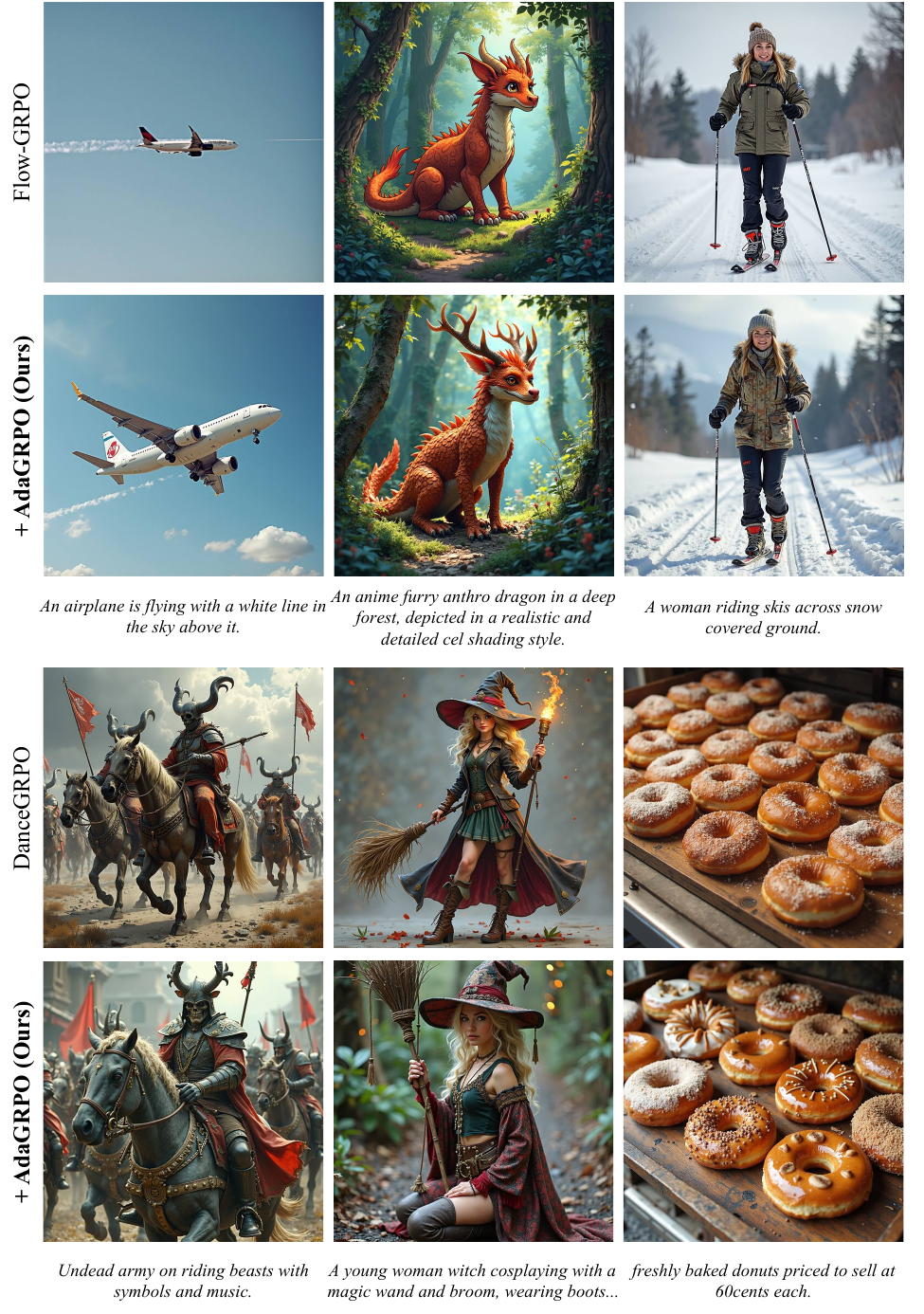}
    \vspace{-1em}
    \caption{
        \textbf{Additional Comparison Results on HPS-v3. (1/2)}
        }
    \label{fig:hpsv3-1}
\end{figure}

\begin{figure}[h]
    \centering
    \includegraphics[width=1.0\linewidth]{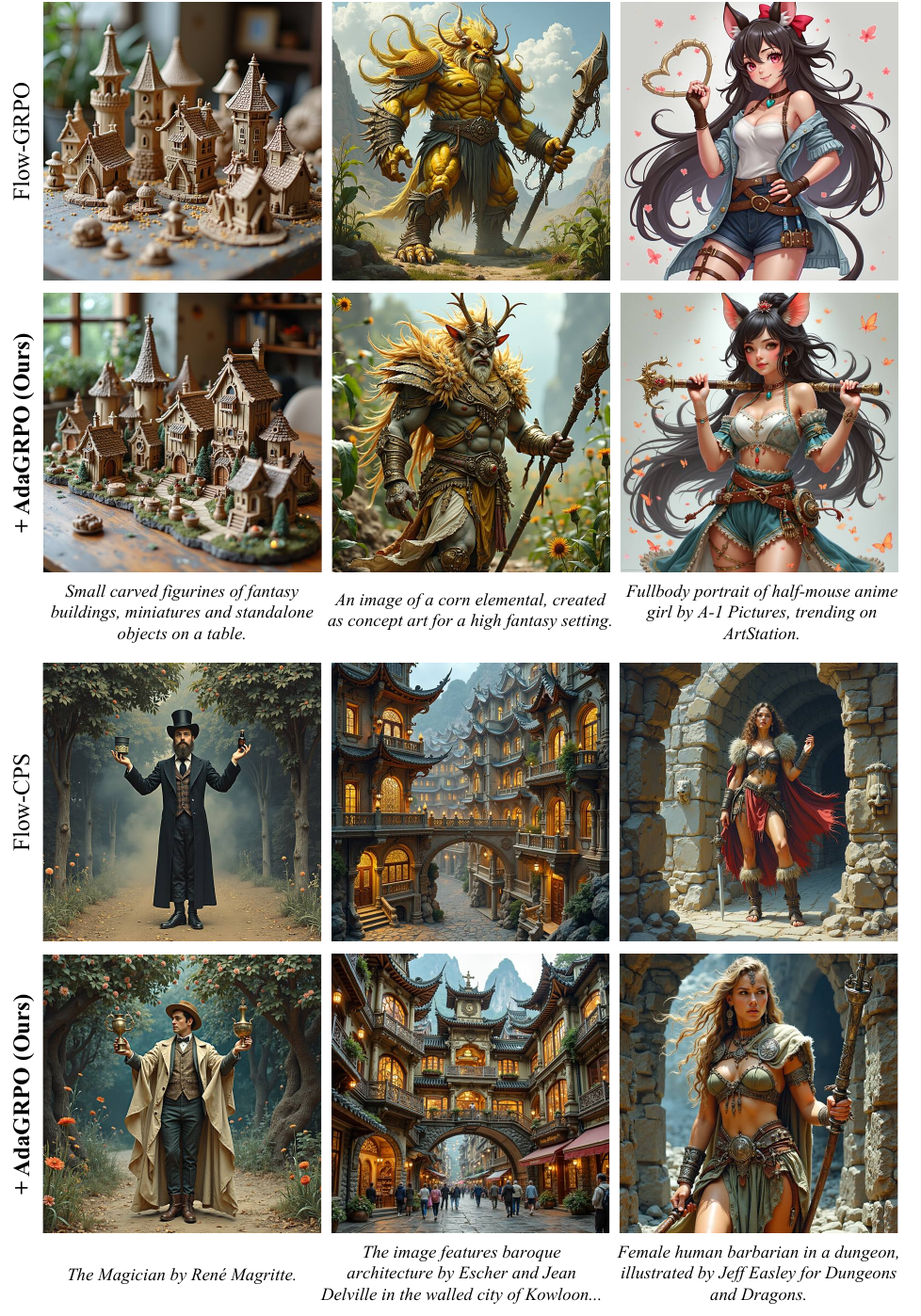}
    \vspace{-1em}
    \caption{
        \textbf{Additional Comparison Results on HPS-v3. (2/2)}
        }
    \label{fig:hpsv3-2}
\end{figure}

\begin{figure}[h]
    \centering
    \includegraphics[width=1.0\linewidth]{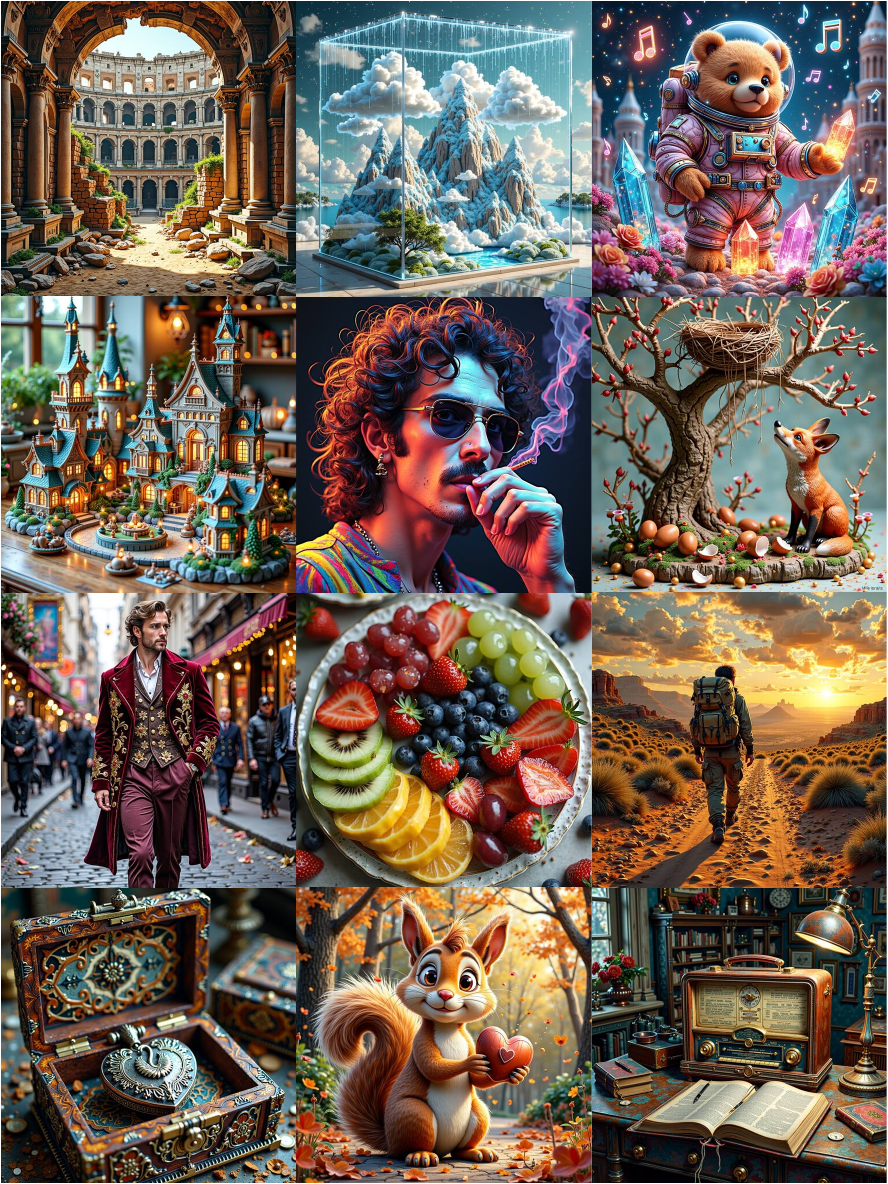}
    \vspace{-1em}
    \caption{
        \textbf{Additional Visual Samples of AdaGRPO. (1/4)}
        }
    \label{fig:gallery-1}
\end{figure}

\begin{figure}[h]
    \centering
    \includegraphics[width=1.0\linewidth]{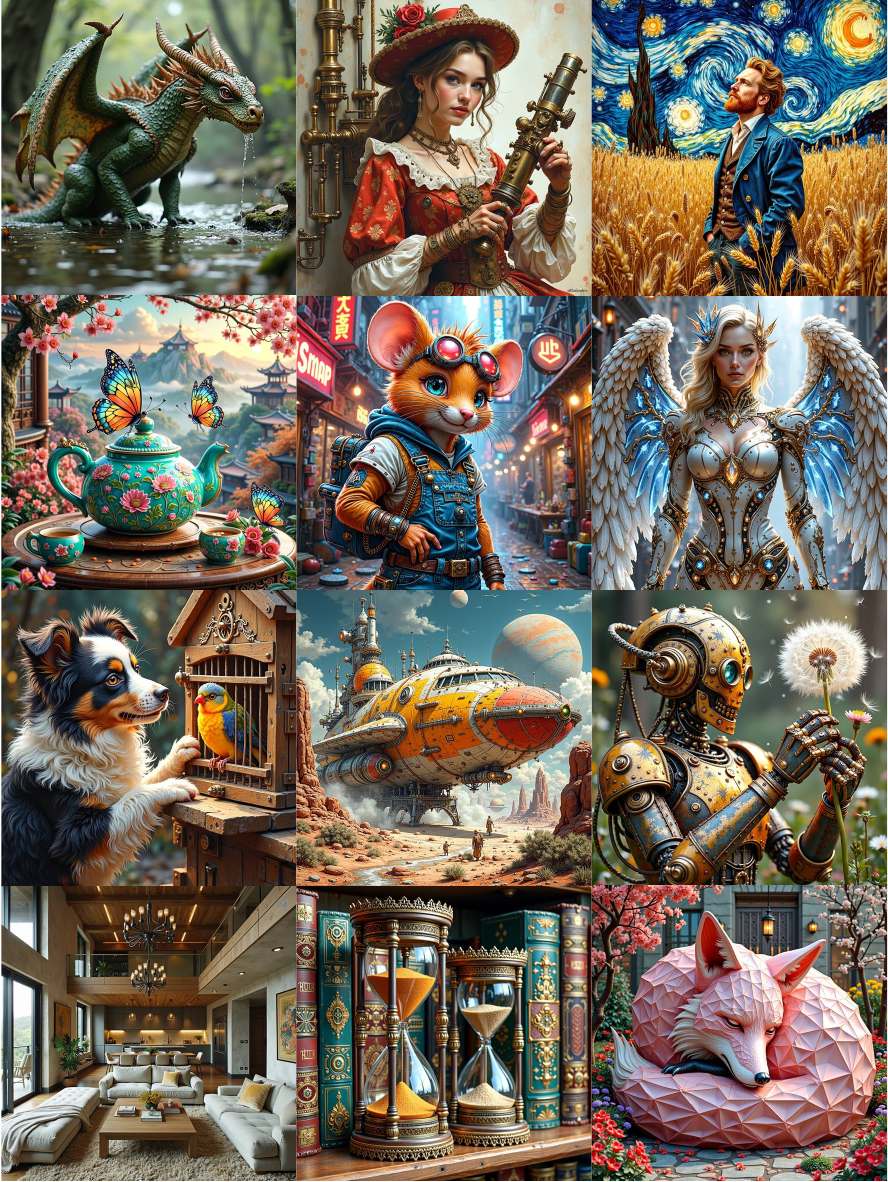}
    \vspace{-1em}
    \caption{
        \textbf{Additional Visual Samples of AdaGRPO. (2/4)}
        }
    \label{fig:gallery-2}
\end{figure}

\begin{figure}[h]
    \centering
    \includegraphics[width=1.0\linewidth]{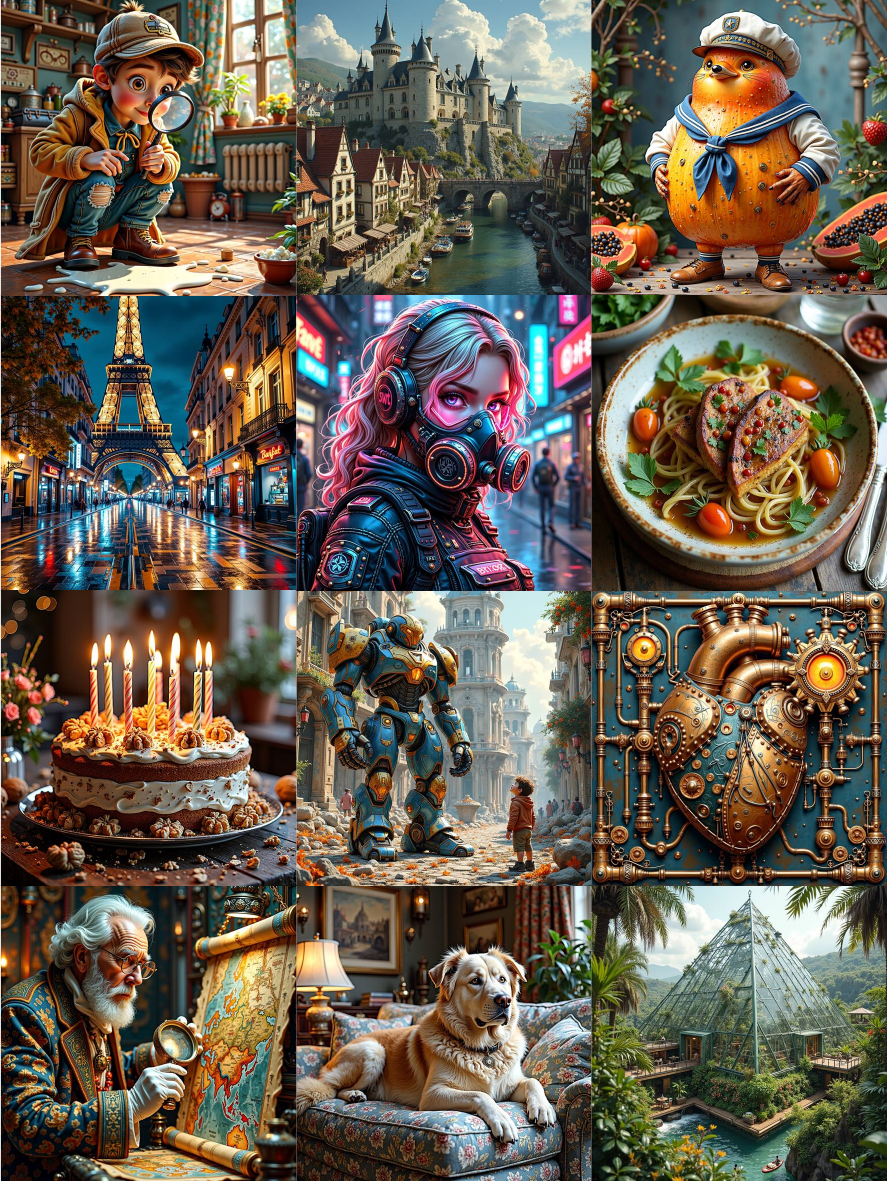}
    \vspace{-1em}
    \caption{
        \textbf{Additional Visual Samples of AdaGRPO. (3/4)}
        }
    \label{fig:gallery-3}
\end{figure}

\begin{figure}[h]
    \centering
    \includegraphics[width=1.0\linewidth]{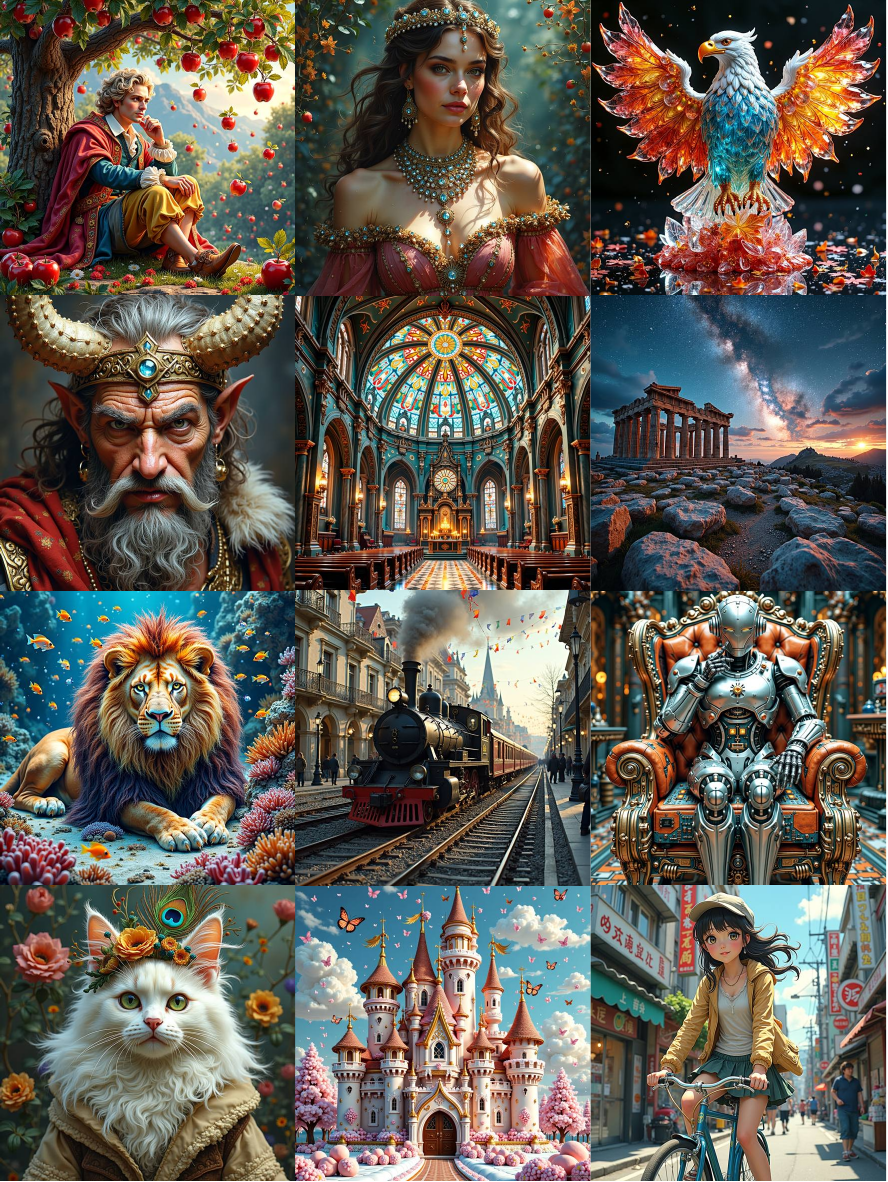}
    \vspace{-1em}
    \caption{
        \textbf{Additional Visual Samples of AdaGRPO. (4/4)}
        }
    \label{fig:gallery-4}
\end{figure}

\begin{figure}[h]
    \centering
    \includegraphics[width=1.0\linewidth]{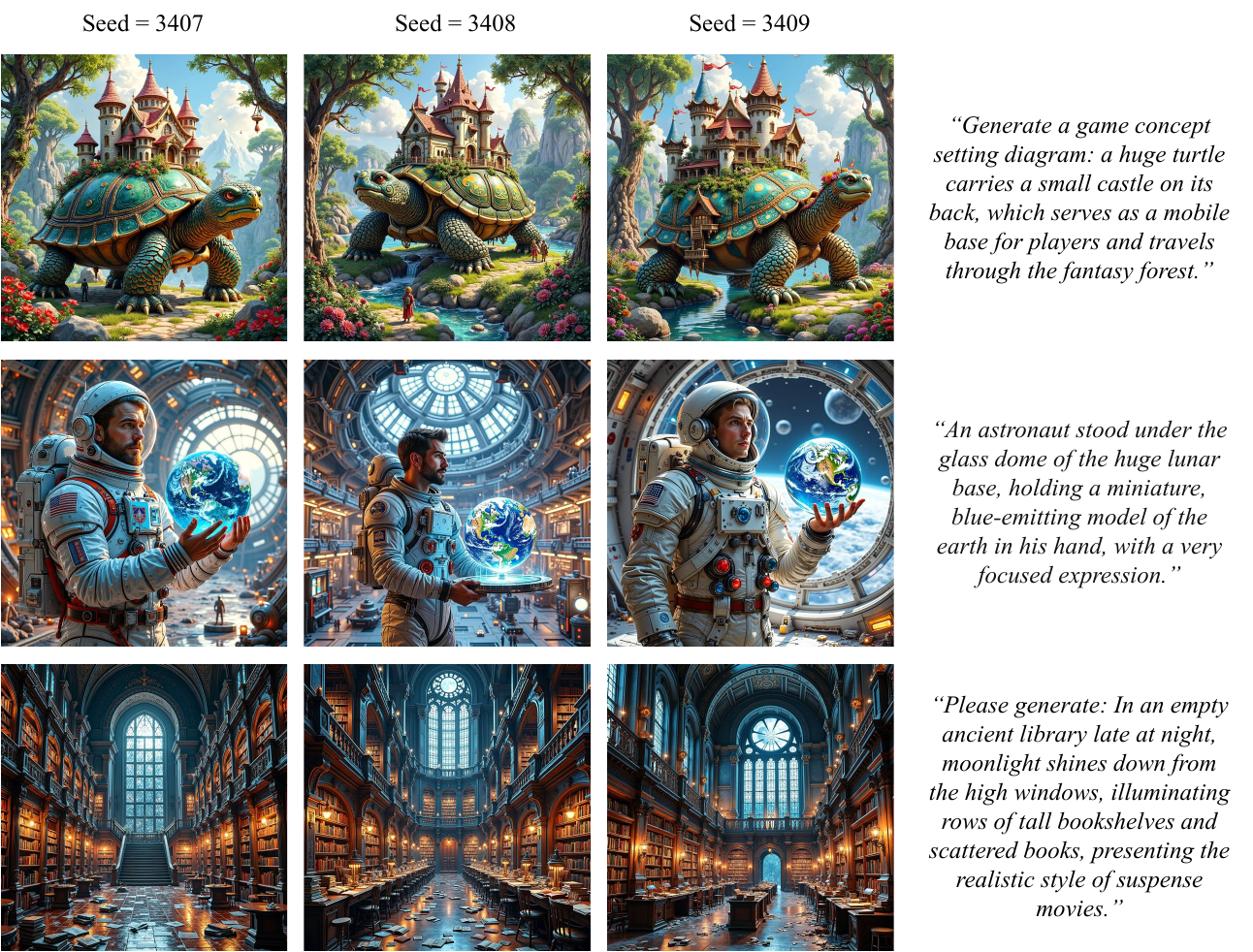}
    \vspace{-1.5em}
    \caption{
        \textbf{Results using same prompts and different seeds. (HPS-v2)}
        }
    \label{fig:seed-hpsv2}
\end{figure}

\begin{figure}[h]
    \centering
    \includegraphics[width=1.0\linewidth]{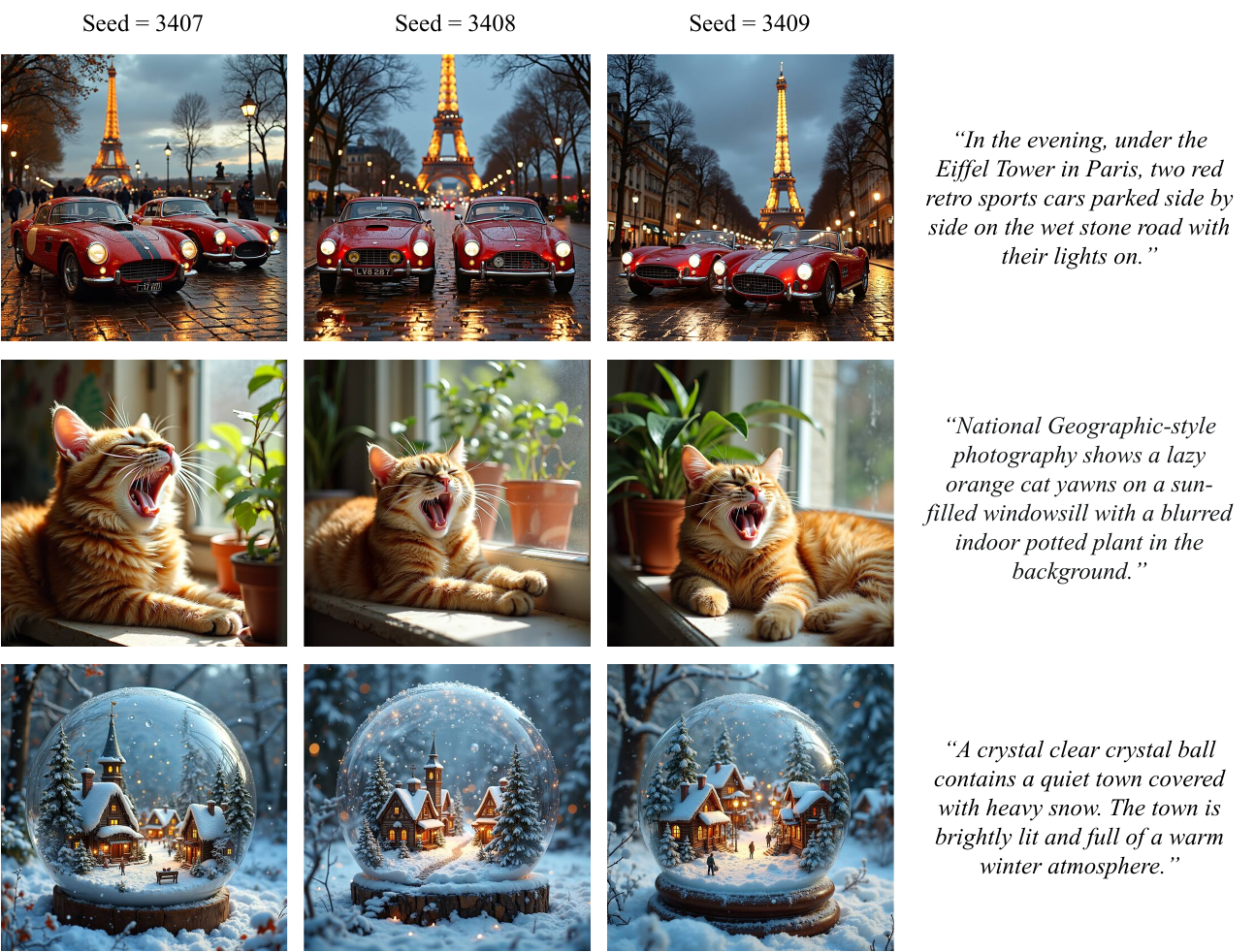}
    \vspace{-1.5em}
    \caption{
        \textbf{Results using same prompts and different seeds. (HPS-v3)}
        }
    \label{fig:seed-hpsv3}
\end{figure}

In this section, we provide additional qualitative comparisons between the proposed AdaGRPO and baseline methods, as shown in Fig.~\ref{fig:hpsv2-1}, Fig.~\ref{fig:hpsv2-2}, Fig.~\ref{fig:hpsv3-1}, and Fig.~\ref{fig:hpsv3-2}. 
We also present more visual results of AdaGRPO in Fig.~\ref{fig:gallery-1}, Fig.~\ref{fig:gallery-2}, Fig.~\ref{fig:gallery-3}, and Fig.~\ref{fig:gallery-4}, as well as generated results obtained using the same prompts but different random seeds in Fig.~\ref{fig:seed-hpsv2} and Fig.~\ref{fig:seed-hpsv3}.

\section{Text Prompts}
\label{sec:prompt}
Text prompts used to generate images in this paper are listed in Tab.~\ref{tab:prompt1}, Tab.~\ref{tab:prompt2} and Tab.~\ref{tab:prompt3}.

\section{Limitation and Discussion}
\label{sec:add-limitation}
While AdaGRPO demonstrates superior performance and training stability, it faces certain constraints. Similar to dynamic data sampling strategies in LLM alignment~\citep{yu2025dapo,bae2026online}, our online prompt filtering mechanism inevitably introduces some computational overhead. However, given the relatively modest VRAM requirements of T2I generation, the deterministic ODE samplings for all candidate prompts within a batch can be efficiently executed in parallel. Consequently, in practice, this profiling process increases the per-iteration training time by merely $\sim$20\%. Future work could focus on exploring more efficient online prompt filtering strategies for flow-based GRPO, aiming to swiftly identify moderate prompts tailored to the model's evolving capabilities. One potential avenue is to employ a low-bit quantized model (e.g., \texttt{int4}) during the filtering phase, reserving the full-precision model (e.g., \texttt{fp16}) exclusively for the subsequent SDE rollouts.

\section{Ethical Statement}
\label{sec:add-ethical}

Throughout the development of this work, we remain steadfast in our dedication to strict moral principles and the responsible advancement of generative AI technologies. To the best of our knowledge, the datasets, algorithmic designs, and downstream applications involved in this study do not introduce any societal risks or ethical hazards. Furthermore, all empirical evaluations and data processing procedures were conducted strictly following widely recognized community norms, guaranteeing the absolute transparency and scientific integrity of our findings.

\section{Reproducibility Statement}
\label{sec:add-reproducibility}

Driven by a strong commitment to open science, we strive to make our experimental results fully verifiable and accessible to the broader academic community. To this end, the complete source code and training scripts of AdaGRPO will be made publicly available. We sincerely hope that these open-source assets will serve as a robust baseline for subsequent studies focusing on reinforcement learning and flow model alignment, ultimately catalyzing further algorithmic breakthroughs and propelling the collective advancement of the field.

\section{Declaration on LLM Usage}
\label{sec:add-llm}

In this paper, we use LLMs only for minor language polishing.

\newpage

\newpage

\clearpage
\begin{table}[!t]
    \centering
    \caption{The image generation prompts for each figure are listed sequentially, following the order from left to right and top to bottom. (Table 1/3)}
     \label{tab:prompt1}
     \resizebox{\textwidth}{!}{%
        \begin{tabular}{>{\arraybackslash}p{2cm} >{\arraybackslash}p{15cm}}
            \toprule
           \textbf{\centering{Figure}}&\textbf{\centering Text Prompt}\\
            \midrule
            \multirow{21}{*}{Figure.~\ref{fig:teaser}} &{There was an open retro wooden jewelry box with an exquisite sapphire necklace lying quietly inside, shining with a glimmer.} \\
 &\cellcolor{color4} Very ornate bedroom with a chandelier over the bed. \\      
  & A spaceship in an empty landscape.  \\     
   &\cellcolor{color4} A Japanese castle landscape painting trending on Artstation. \\  
  & A giant panda dressed in the costume of an ancient Egyptian pharaoh sits upright on a golden throne in the dense bamboo forest. \\ 
  &\cellcolor{color4} A photorealistic, ultra-detailed macro shot of an exquisite artisanal chocolate cake resting on an elegant vintage silver cake stand. The glossy, rich chocolate ganache reflects soft, warm ambient lighting. In the center of the cake, the text ``Adaptive GRPO'' is flawlessly written in elegant white vanilla cream piping. The cake is intricately decorated with delicate edible gold leaf flakes, fresh plump raspberries, and dark chocolate curls. The background is a sophisticated, dimly lit fine dining restaurant with beautiful warm ambient bokeh blurring out crystal glasses and glowing candles. Cinematic lighting, highly appetizing, high-end food photography, 8k resolution. \\
  & A Monet portrait. \\
  &\cellcolor{color4} A turtle with feathered wings is not in the air because it cannot fly, but lies leisurely on the clouds to rest. \\
  & On the street of the future of cyberpunk-style Tokyo, a woman wearing VR glasses controls the holographic koi floating in front of her through the air. \\
  &\cellcolor{color4} A crystal wall clock in the shape of an ancient Roman Colosseum. \\
  & Please generate a documentary black and white photo of a tiny astronaut standing next to the ancient ruins of a giant planet. \\
  \midrule
    \multirow{3}{*}{Figure.~\ref{fig:observation}}
    & A bear walks through a group of bushes with a plant in its mouth.\\
    &\cellcolor{color4} A hand-drawn cute gnome holding a pumpkin in an autumn disguise, portrayed in a detailed close-up of the face with warm lighting and high detail. \\ 
            \midrule
            \multirow{6}{*}{Figure.~\ref{fig:hpsv2}} &{People standing in the grass playing with a frisbee.}\\
 &\cellcolor{color4} Two young ladies seated with several other people at a dinner table.  \\     
   & Hooded figure standing over a ruined city with red haze and a grin.\\  
  &\cellcolor{color4} Portrait of an anime princess in white and golden clothes. \\ 
  & Side of a street, where there is a fire hydrant and a mirror showing the street. \\
  &\cellcolor{color4} Side-view blue-ice sneaker inspired by Spiderman created by Weta FX. \\
  \midrule
    \multirow{9}{*}{Figure.~\ref{fig:hpsv3}} &{A cute anime schoolgirl with a sad face submerged in dark pink and blue water, portrayed in an oil painting style.}\\
 &\cellcolor{color4} A warrior in glowing azure plate armor stands in a doorway to hell sliced by iridescent glass cracks, with crimson clouds and an art deco palace backdrop.  \\     
   & A man with his finger held up to his nose. \\  
  &\cellcolor{color4} A photorealistic image from a furry fandom convention set in a biopunk era after the genetic revolution and quantum singularity. \\ 
  & A phantom airship. \\
  &\cellcolor{color4} A portrait painting of Leighann Vail. \\
    \midrule
    \multirow{6}{*}{Figure.~\ref{fig:hpsv2-1}} &{A 3D rendering of anime schoolgirls with a sad expression underwater, surrounded by dramatic lighting.}\\
 &\cellcolor{color4} A couple of men are standing outside their car watching sheep cross a road. \\     
   & A cute little anthropomorphic Tropical fish knight wearing a cape and a crown in short, pale blue armor. \\  
  &\cellcolor{color4} A detailed and realistic fantasy Proto-Slavic skinny red troll creature. \\ 
  & A man walking on the bea with his surfboard. \\
  &\cellcolor{color4} A fruit basket on a kitchen table with a Studio Ghibli reference. \\
  \midrule
    \multirow{8}{*}{Figure.~\ref{fig:hpsv2-2}} &{A Wojak looking over a sea of memes from a cliff on 4chan.}\\
 &\cellcolor{color4} A steampunk pocketwatch owl is trapped inside a glass jar buried in sand, surrounded by an hourglass and swirling mist. \\     
   & American cowboy with a scruffy appearance in a retrofuturistic style, inspired by the animations of Studio Ghibli. \\  
  &\cellcolor{color4} A person flying a kite while standing in the grass. \\ 
  & An image of an emo with dark brown hair in a messy pixie cut, large entirely-black eyes, wearing black clothing and boots. \\
  &\cellcolor{color4} Album art of a hand holding a balloon emerging from the water against a red sky. \\
  \midrule
    \multirow{7}{*}{Figure.~\ref{fig:hpsv3-1}} &{An airplane is flying with a white line in the sky above it.}\\
 &\cellcolor{color4} An anime furry anthro dragon in a deep forest, depicted in a realistic and detailed cel shading style. \\     
   & A woman riding skis across snow covered ground. \\  
  &\cellcolor{color4} Undead army on riding beasts with symbols and music. \\ 
  & A young woman witch cosplaying with a magic wand and broom, wearing boots, and posing in a full body shot with a detailed face. \\
  &\cellcolor{color4} freshly baked donuts priced to sell at 60cents each. \\
\bottomrule
        \end{tabular}
    }
\end{table}

\clearpage
\begin{table}[!t]
    \centering
    \caption{The image generation prompts for each figure are listed sequentially, following the order from left to right and top to bottom. (Table 2/3)}
     \label{tab:prompt2}
     \resizebox{\textwidth}{!}{%
        \begin{tabular}{>{\arraybackslash}p{2cm} >{\arraybackslash}p{15cm}}
            \toprule
           \textbf{\centering{Figure}}&\textbf{\centering Text Prompt}\\
  \midrule
    \multirow{7}{*}{Figure.~\ref{fig:hpsv3-2}} &{Small carved figurines of fantasy buildings, miniatures and standalone objects on a table.}\\
 &\cellcolor{color4} An image of a corn elemental, created as concept art for a high fantasy setting. \\     
   & Fullbody portrait of half-mouse anime girl by A-1 Pictures, trending on ArtStation. \\  
  &\cellcolor{color4} The Magician by René Magritte. \\ 
  & The image features baroque architecture by Escher and Jean Delville in the walled city of Kowloon, lit with golden lighting and displaying ornate details. \\
  &\cellcolor{color4} Female human barbarian in a dungeon, illustrated by Jeff Easley for Dungeons and Dragons. \\
            \midrule
            \multirow{18}{*}{Figure.~\ref{fig:gallery-1}} &{From the interior perspective of the ancient Roman Colosseum, sunlight shines in through the dilapidated roof structure, illuminating the sand and passages below.} \\
 &\cellcolor{color4} A miniature snowy mountain wonderland enclosed in glass. \\      
  & Please design a playback scene for a music App. A cartoon bear wearing a spacesuit floats to touch the musical symbols made of crystals. The overall style is a glass imitation.  \\     
   &\cellcolor{color4} Small carved figurines of fantasy buildings, miniatures and standalone objects on a table. \\  
  & A portrait of Frank Zappa smoking, with vivid neon colors, by various artists. \\ 
  &\cellcolor{color4} In a clay sculpture, a hungry fox looks up at an empty bird's nest on a high branch, with broken eggshells scattered on the ground below. \\
  & a man walking alone down the street in a velvet jacket. \\
  &\cellcolor{color4} A plate topped with lots of different kinds of fruit. \\
  & In the golden desert at dusk, a traveler carrying a huge backpack is struggling to walk into the distance, trailing a long shadow behind him. \\
  &\cellcolor{color4} An ornate box with a heart-shaped jewel inside. \\
  & Please generate a cartoon-style picture of a squirrel holding a heart-shaped nut in his hand, with a confused expression, but it does not open its mouth to eat it. \\
  &\cellcolor{color4} An old-fashioned radio was placed on the dusty desk. It didn't play any sound. The diary next to it was open, and the atmosphere was suspenseful. \\
  \midrule
\multirow{16}{*}{Figure.~\ref{fig:gallery-2}} &{A dragon standing in a forest, drinking river water.} \\
 &\cellcolor{color4} A steampunk picture of a lady in a classic dress operating a complex brass mechanism. \\      
  & Vincent Van Gogh stood in the golden wheat field. He looked up at the stars with melancholy and fanaticism on his face, with a cinematic texture like an oil painting.  \\     
   &\cellcolor{color4} A celadon teapot around with colorful butterflies among the hazy distant mountains. It has a strong national style and ink painting style. \\  
  & A ginger haired mouse mechanic in blue overalls in a cyberpunk scene with neon slums in the background. \\ 
  &\cellcolor{color4} A photo of a mechanical angel woman with crystal wings, in the sci-fi style of Stefan Kostic, created by Stanley Lau and Artgerm. \\
  & A curious dog greets a colorful parrot leaning out of its wooden cage. \\
  &\cellcolor{color4} A massive and brightly colored spacecraft in a deserted landscape, depicted in retro 1960s sci-fi art. \\
  & Generating picture: A steampunk-style robot is stretching out its brass fingers and carefully touching a fragile dandelion. \\
  &\cellcolor{color4} A great room with the living area in the foreground, dining table behind it and kitchen in the very back.  \\
  & There are two hourglasses on the bookshelf. \\
  &\cellcolor{color4} A geometric pink fox resting among blooming flowers. \\
   \midrule
\multirow{18}{*}{Figure.~\ref{fig:gallery-3}} &{In Pixar animation style, a young detective holding a magnifying glass squatted on the ground and carefully observed a pool of spilled milk.} \\
 &\cellcolor{color4} a castle is in the middle of a eurpean city. \\      
  & a papaya fruit dressed as a sailor.  \\     
   &\cellcolor{color4} Under the Eiffel Tower in Paris at night, a quiet street is arranged as a product launch scene. The ground is wet and reflects light, and the overall futuristic technological style is adopted. \\  
  & A key visual of a young female swat officer with a neon futuristic gas mask in a cyberpunk setting. \\ 
  &\cellcolor{color4} a plate that has some kind of food on it. \\
  & A lighted birthday cake with chunks of walnuts. \\
  &\cellcolor{color4} In the ruined city, a little boy looked up in surprise at the huge, still robot in front of him, with a cinematic lens and a realistic style. \\
  & A bronze steampunk mechanical heart, intertwined with gears and pipes, exuding a faint orange glow, surrealist oil painting. \\
  &\cellcolor{color4} An elderly historian wearing white cotton gloves carefully examined a yellowed sheepskin scroll map with a magnifying glass, with a solemn expression. \\
  & A large dog laying on a couch in a room. \\
  &\cellcolor{color4} A huge glass greenhouse shaped like the Great Pyramid of Giza contains a complete and miniature Amazon rainforest ecosystem, and the overall surrealist style. \\
\bottomrule
        \end{tabular}
    }
\end{table}

\clearpage
\begin{table}[!t]
    \centering
    \caption{The image generation prompts for each figure are listed sequentially, following the order from left to right and top to bottom. (Table 3/3)}
     \label{tab:prompt3}
     \resizebox{\textwidth}{!}{%
        \begin{tabular}{>{\arraybackslash}p{2cm} >{\arraybackslash}p{15cm}}
            \toprule
           \textbf{\centering{Figure}}&\textbf{\centering Text Prompt}\\
  \midrule
\multirow{20}{*}{Figure.~\ref{fig:gallery-4}} &{Generating picture: depicts the famous scene of physicist Newton sitting under an apple tree, falling into thought after being hit by a falling apple.} \\
 &\cellcolor{color4} An oil painting portrait of a beautiful dryad wearing an ombre velvet gown, with long hair and a tiara, adorned with dozens of jeweled necklaces, and illuminated with dramatic cinematic lighting. \\      
  & An eagle carved out of crystal has huge, colorful maple leaves that spread its wings on the base and want to fly.  \\     
   &\cellcolor{color4} Oil painting portrait of demon king with gazing eyes, art by John Howe, Keith Parkinson, and Larry Elmore, featured on ArtStation and CGSociety. \\  
  & Generated image: Inside an ancient church, the dome is composed of countless stained glass sheets, and the picture presents a solemn cinematic realistic style. \\ 
  &\cellcolor{color4} Close-up view of ancient Greek ruins set against a colourful, starry night sky creating a mystical atmosphere. \\
  & A lion lies quietly on the bottom of the sea. \\
  &\cellcolor{color4} A train that is going by a building. \\
  & A regal robot sits thoughtfully on the throne. \\
  &\cellcolor{color4} A white Persian cat wearing a peacock feather headdress and surrounded by flowers, in a magical realism painting. \\
  & Please draw a castle made of marshmallows, with melted chocolate on top of its towers, and the overall sweet and dreamy style. \\
  &\cellcolor{color4} An anime girl is riding a bicycle in Akihabara, resembling the style seen in Studio Ghibli films, and the depiction is detailed. \\
  \midrule
\multirow{6}{*}{Figure.~\ref{fig:seed-hpsv2}} &{Generate a game concept setting diagram: a huge turtle carries a small castle on its back, which serves as a mobile base for players and travels through the fantasy forest.} \\
 &\cellcolor{color4} An astronaut stood under the glass dome of the huge lunar base, holding a miniature, blue-emitting model of the earth in his hand, with a very focused expression. \\      
  & Please generate: In an empty ancient library late at night, moonlight shines down from the high windows, illuminating rows of tall bookshelves and scattered books, presenting the realistic style of suspense movies.  \\     
  \midrule
\multirow{6}{*}{Figure.~\ref{fig:seed-hpsv3}} &{In the evening, under the Eiffel Tower in Paris, two red retro sports cars parked side by side on the wet stone road with their lights on.} \\
 &\cellcolor{color4} National Geographic-style photography shows a lazy orange cat yawns on a sun-filled windowsill with a blurred indoor potted plant in the background. \\      
  & A crystal clear crystal ball contains a quiet town covered with heavy snow. The town is brightly lit and full of a warm winter atmosphere.  \\     
\bottomrule
        \end{tabular}
    }
\end{table}

\end{document}

%% file: math_commands.tex
\usepackage{amsmath,amsfonts,bm}

\def\eqref#1{equation~\ref{#1}}

\def\1{\bm{1}}

\DeclareMathAlphabet{\mathsfit}{\encodingdefault}{\sfdefault}{m}{sl}
\SetMathAlphabet{\mathsfit}{bold}{\encodingdefault}{\sfdefault}{bx}{n}



%% file: tabs/quant.tex
\begin{table}[t]
\centering
\caption{Quantitative comparison across different settings and frameworks. \textbf{Bold} values indicate the best result within each pair. \textcolor{blue}{Shaded} rows denote results with our AdaGRPO.
UR-v2-A, UR-v2-C and UR-v2-S represent the Alignment, Coherence and Style dimensions of UR-v2, respectively.
}
\resizebox{0.9\textwidth}{!}{%
\begin{tabular}{cl cccc cccc}
\toprule
\textbf{Reward Model} & \textbf{Method} & \textbf{HPS-v2}$\uparrow$ & \textbf{HPS-v3}$\uparrow$ & \textbf{UR-v2-A}$\uparrow$ & \textbf{UR-v2-C}$\uparrow$ & \textbf{UR-v2-S}$\uparrow$ & \textbf{CLIP}$\uparrow$ & \textbf{IR}$\uparrow$ & \textbf{UR-v1}$\uparrow$ \\
\midrule
/ & Flux.1-dev & 0.3065 & 13.2803 & 3.2550 & 3.6547 & 3.2333 & 0.3901 & 1.0546 & 3.6006 \\
\midrule
\multirow{6}{*}{\textbf{HPS-v2}}
& Flow-GRPO
& 0.3463 & 14.4374 & \textbf{3.2293} & 3.6236 & 3.3395
& \textbf{0.3754} & 1.2435 & 3.5167 \\
\rowcolor{oursrow}
\cellcolor{white} & + AdaGRPO
& \textbf{0.3558} & \textbf{14.8870} & 3.2222 & \textbf{3.6595} & \textbf{3.3840}
& 0.3676 & \textbf{1.2645} & \textbf{3.5285} \\
\addlinespace
& DanceGRPO
& 0.3430 & 14.4687 & \textbf{3.2291} & 3.6253 & 3.3089
& \textbf{0.3724} & 1.2178 & 3.5129 \\
\rowcolor{oursrow}
\cellcolor{white} & + AdaGRPO
& \textbf{0.3516} & \textbf{14.5155} & 3.2209 & \textbf{3.6584} & \textbf{3.3891}
& 0.3685 & \textbf{1.2539} & \textbf{3.5355} \\
\addlinespace
& Flow-CPS
& 0.3492 & 14.8044 & \textbf{3.2387} & 3.6660 & 3.3679
& \textbf{0.3745} & 1.2561 & 3.5276 \\
\rowcolor{oursrow}
\cellcolor{white} & + AdaGRPO
& \textbf{0.3579} & \textbf{14.9521} & 3.2264 & \textbf{3.6676} & \textbf{3.3722}
& 0.3691 & \textbf{1.2575} & \textbf{3.5537} \\
\midrule
\multirow{6}{*}{\textbf{HPS-v3}}
& Flow-GRPO
& 0.3272 & 14.7143 & \textbf{3.2300} & 3.6728 & 3.2836
& \textbf{0.3787} & 1.1475 & \textbf{3.5821} \\
\rowcolor{oursrow}
\cellcolor{white} & + AdaGRPO
& \textbf{0.3319} & \textbf{15.0982} & 3.2018 & \textbf{3.6774} & \textbf{3.3154}
& 0.3699 & \textbf{1.1490} & 3.5512 \\
\addlinespace
& DanceGRPO
& 0.3275 & 14.6675 & \textbf{3.2345} & 3.6450 & 3.2843
& \textbf{0.3813} & 1.1431 & \textbf{3.5460} \\
\rowcolor{oursrow}
\cellcolor{white} & + AdaGRPO
& \textbf{0.3301} & \textbf{14.9501} & 3.2266 & \textbf{3.6724} & \textbf{3.3052}
& 0.3780 & \textbf{1.1487} & 3.5376 \\
\addlinespace
& Flow-CPS
& 0.3291 & 14.7578 & \textbf{3.2312} & 3.6550 & 3.2786
& \textbf{0.3783} & \textbf{1.1598} & \textbf{3.5375} \\
\rowcolor{oursrow}
\cellcolor{white} & + AdaGRPO
& \textbf{0.3311} & \textbf{15.1093} & 3.2169 & \textbf{3.6823} & \textbf{3.3133}
& 0.3734 & 1.1548 & 3.5193 \\
\midrule
\multirow{6}{*}{\textbf{HPS-v3 + CLIP}}
& Flow-GRPO
& 0.3250 & 14.2459 & \textbf{3.2605} & 3.6231 & 3.2487
& 0.3938 & 1.1988 & 3.6260 \\
\rowcolor{oursrow}
\cellcolor{white} & + AdaGRPO
& \textbf{0.3288} & \textbf{14.6114} & 3.2566 & \textbf{3.6418} & \textbf{3.2511}
& \textbf{0.3952} & \textbf{1.2018} & \textbf{3.6340} \\
\addlinespace
& DanceGRPO
& 0.3189 & 13.8573 & 3.2483 & 3.6294 & 3.2499
& 0.3914 & 1.1406 & 3.5859 \\
\rowcolor{oursrow}
\cellcolor{white} & + AdaGRPO
& \textbf{0.3246} & \textbf{14.3156} & \textbf{3.2514} & \textbf{3.6357} & \textbf{3.2500}
& \textbf{0.3924} & \textbf{1.2216} & \textbf{3.6203} \\
\addlinespace
& Flow-CPS
& 0.3252 & 14.3248 & 3.2608 & 3.6180 & 3.2358
& 0.3926 & 1.2222 & 3.6339 \\
\rowcolor{oursrow}
\cellcolor{white} & + AdaGRPO
& \textbf{0.3286} & \textbf{14.6044} & \textbf{3.2642} & \textbf{3.6338} & \textbf{3.2475}
& \textbf{0.3942} & \textbf{1.2300} & \textbf{3.6419} \\
\bottomrule
\end{tabular}%
}
\vspace{-1em}
\label{tab:comparison}
\end{table} 

%% file: tabs/unigen.tex
\begin{table}[h]
\centering
\caption{Quantitative comparison on UniGenBench (Training Setting: HPS-v3 + CLIP). \textbf{Bold} values indicate the best result within each pair. \textcolor{blue}{Shaded} rows denote results with our AdaGRPO. 
}
\resizebox{0.9\textwidth}{!}{%
\begin{tabular}{l c|cccccccccc}
\toprule
\textbf{Method} & \textbf{Overall}$\uparrow$ & \textbf{Style}$\uparrow$ & \textbf{Know.}$\uparrow$ & \textbf{Attri.}$\uparrow$ & \textbf{Act.}$\uparrow$ & \textbf{Rel.}$\uparrow$ & \textbf{Comp.}$\uparrow$ & \textbf{Gram.}$\uparrow$ & \textbf{Reason.}$\uparrow$ & \textbf{Layout}$\uparrow$ & \textbf{Text}$\uparrow$ \\
\midrule
Flux.1-dev & 59.86 & 84.60 & 85.76 & 64.32 & 61.22 & 66.37 & 46.26 & 58.69 & 27.06 & 70.71 & 33.62 \\
\midrule
Flow-GRPO & 62.65 & 85.80 & 89.08 & 70.19 & 66.06 & 69.67 & 51.03 & \textbf{60.70} & 29.36 & 73.88 & \textbf{30.75} \\
\rowcolor{oursrow}
+ AdaGRPO & \textbf{63.24} & \textbf{86.40} & \textbf{89.56} & \textbf{70.73} & \textbf{66.83} & \textbf{71.70} & \textbf{51.16} & 60.29 & \textbf{30.28} & \textbf{75.00} & 30.46 \\
\addlinespace
DanceGRPO & 61.15 & 85.00 & 88.29 & 68.48 & 61.12 & \textbf{70.05} & 48.20 & 59.36 & 28.90 & 71.64 & \textbf{30.46} \\
\rowcolor{oursrow}
+ AdaGRPO & \textbf{62.41} & \textbf{86.50} & \textbf{90.51} & \textbf{69.02} & \textbf{62.93} & 69.42 & \textbf{51.68} & \textbf{60.56} & \textbf{30.50} & \textbf{73.13} & 29.89 \\
\addlinespace
Flow-CPS & 62.65 & 86.10 & 90.03 & 69.66 & \textbf{65.87} & 71.32 & 52.19 & \textbf{61.36} & 29.13 & 73.51 & 27.30 \\
\rowcolor{oursrow}
+ AdaGRPO & \textbf{63.31} & \textbf{86.50} & \textbf{90.66} & \textbf{69.98} & 65.11 & \textbf{72.84} & \textbf{52.71} & 60.70 & \textbf{29.82} & \textbf{74.07} & \textbf{30.75} \\
\bottomrule
\end{tabular}%
}
\vspace{-1em}
\label{tab:unigenbench}
\end{table}

%% file: tabs/ablation_right.tex
\begin{wraptable}{R}{0.55\textwidth}
\vspace{-2.25em}
\centering
\caption{Ablation study. \textbf{Bold} and \underline{underlined} indicate the best and second-best results, respectively.}
\resizebox{\linewidth}{!}{%
\begin{tabular}{cc ccc}
\toprule
\textbf{Components} & \textbf{Choices} & \textbf{HPS-v2}$\uparrow$ & \textbf{HPS-v3}$\uparrow$ & \textbf{UR-v2 (averaged)}$\uparrow$ \\
\midrule
\multirow{5}{*}{(a) Momentum Coefficient $\alpha$}
& $0.9$ & 0.3490 & 14.7034 & 3.4107 \\
& $0.8$ & 0.3493 & 14.6675 & \underline{3.4212} \\
& $0.7$ & 0.3499 & 14.6815 & \textbf{3.4241} \\
& \cellcolor{oursrow}$\mathbf{0.6}$ & \cellcolor{oursrow}\textbf{0.3507} & \cellcolor{oursrow}\textbf{14.7761} & \cellcolor{oursrow}3.4197 \\
& $0.5$ & \underline{0.3503} & \underline{14.7340} & 3.4175 \\
\midrule
\multirow{3}{*}{(b) Candidate Batch Size $B$}
& $5$ & 0.3480 & 14.6277 & 3.4135 \\
& \cellcolor{oursrow}$\mathbf{10}$ & \cellcolor{oursrow}\underline{0.3507} & \cellcolor{oursrow}\underline{14.7761} & \cellcolor{oursrow}\underline{3.4197} \\
& $20$ & \textbf{0.3510} & \textbf{14.8026} & \textbf{3.4199} \\
\midrule
\multirow{2}{*}{(c) Cross-Level Adv Fusion}
& w/o & 0.3507 & 14.7761 & 3.4197 \\
& \cellcolor{oursrow}\textbf{w} & \cellcolor{oursrow}\textbf{0.3558} & \cellcolor{oursrow}\textbf{14.8870} & \cellcolor{oursrow}\textbf{3.4219} \\
\bottomrule
\end{tabular}%
}
\vspace{-2em}
\label{tab:ablation}
\end{wraptable}

%% file: tabs/hyperparameters.tex
\begin{table}[ht]
\caption{Hyperparameter settings in our experiments.}
\centering
\resizebox{0.6\linewidth}{!}{
\begin{tabular}{lclc}
\toprule
\textbf{Parameter} & \textbf{Value} & \textbf{Parameter} & \textbf{Value} \\
\midrule
Random seed & 42 & Learning rate & $2\times 10^{-6}$ \\
Train batch size & 1 & Weight decay & $1\times 10^{-4}$ \\
Warmup steps & 0 & Mixed precision & \texttt{bfloat16} \\
Dataloader workers & 4 & Max grad norm & 1.0 \\
Eta & 0.7 & Sampler seed & 1223627 \\
Group size & 12 & Scheduler shift  & 3 \\
Sampling steps & 16  & Adv. clip max & 5.0 \\
Init same noise & Yes & Training steps & $\{0,2,4,6\}$ \\
The number of GPUs & 8 &  Clip range  & $1\times 10^{-4}$\\
\bottomrule
\label{tab:hyperparams}
\end{tabular}
}
\end{table}

%% file: tabs/unifiedrewardv2.tex
\begin{table}[h]
\centering
\caption{Quantitative comparison of models trained with UnifiedReward-v2 (UR-v2). \textbf{Bold} values indicate the best result within each pair. \textcolor{blue}{Shaded} rows denote results with our AdaGRPO.
}
\resizebox{\textwidth}{!}{%
\begin{tabular}{cl cccc cccc}
\toprule
\textbf{Reward Model} & \textbf{Method} & \textbf{HPS-v2}$\uparrow$ & \textbf{HPS-v3}$\uparrow$ & \textbf{UR-v2-A}$\uparrow$ & \textbf{UR-v2-C}$\uparrow$ & \textbf{UR-v2-S}$\uparrow$ & \textbf{CLIP}$\uparrow$ & \textbf{IR}$\uparrow$ & \textbf{UR-v1}$\uparrow$ \\
\midrule
/ & Flux.1-dev & 0.3065 & 13.2803 & 3.2550 & 3.6547 & 3.2333 & 0.3901 & 1.0546 & 3.6006 \\
\midrule
\multirow{6}{*}{\textbf{UR-v2}}
& Flow-GRPO
& 0.3173 & 13.8476 & \textbf{3.2390} & 3.6973 & 3.3392
& \textbf{0.3823} & 1.1127 & 3.5376 \\
\rowcolor{oursrow}
\cellcolor{white} & + AdaGRPO
& \textbf{0.3202} & \textbf{13.9199} & 3.2320 & \textbf{3.7358} & \textbf{3.3670}
& 0.3771 & \textbf{1.1159} & \textbf{3.5804} \\
\addlinespace
& DanceGRPO
& 0.3202 & 14.0876 & 3.2336 & 3.6807 & 3.3162
& 0.3802 & 1.0962 & 3.5181 \\
\rowcolor{oursrow}
\cellcolor{white} & + AdaGRPO
& \textbf{0.3210} & \textbf{14.1153} & \textbf{3.2458} & \textbf{3.7058} & \textbf{3.3297}
& \textbf{0.3836} & \textbf{1.1387} & \textbf{3.5980} \\
\addlinespace
& Flow-CPS
& 0.3213 & 14.1655 & \textbf{3.2406} & 3.7146 & 3.3794
& \textbf{0.3786} & 1.1246 & 3.5639 \\
\rowcolor{oursrow}
\cellcolor{white} & + AdaGRPO
& \textbf{0.3265} & \textbf{14.1743} & 3.2382 & \textbf{3.7272} & \textbf{3.3925}
& 0.3774 & \textbf{1.1452} & \textbf{3.5657} \\
\bottomrule
\end{tabular}%
}
\label{tab:urv2_quant}
\end{table}